\documentclass[10pt]{article}

\usepackage[T1]{fontenc}
\usepackage{times}
\usepackage[
    letterpaper,
    left=1in,
    right=1in,
    top=0.78in,
    bottom=0.78in,
    headheight=20pt,
    headsep=16pt,
    footskip=24pt
]{geometry}
\usepackage{microtype}
\usepackage[round,authoryear]{natbib}
\usepackage{fancyhdr}
\usepackage{titlesec}

\usepackage[table]{xcolor}
\usepackage{booktabs}
\usepackage{multirow}
\usepackage{url}
\usepackage{graphicx}
\usepackage{adjustbox}
\usepackage{array}
\usepackage{makecell}
\usepackage{amsmath}
\usepackage{algorithm}
\usepackage{algorithmic}
\usepackage{multicol}
\usepackage{rotating}
\usepackage{xspace}
\usepackage{subcaption}
\usepackage{tabularx}
\usepackage{enumitem}
\usepackage{placeins}
\usepackage{amssymb}
\usepackage{pifont}
\usepackage{wrapfig}
\usepackage{tikz}
\usepackage[most]{tcolorbox}

\definecolor{SFBlue}{HTML}{205F80}
\definecolor{SFLinkBlue}{HTML}{55AEDD}
\definecolor{SFRuleBlue}{HTML}{68AED0}
\definecolor{SFBoxBG}{HTML}{EDF6FA}
\definecolor{SFBoxBorder}{HTML}{BDDDEB}

\usepackage[
    colorlinks=true,
    linkcolor=SFBlue,
    citecolor=SFLinkBlue,
    urlcolor=SFLinkBlue
]{hyperref}
\hypersetup{
    pdftitle={Opera: A Verbal Critic Framework for Long-horizon Coding Agents},
    pdfauthor={Kai Mei, Zhiyuan Hu, Yutong Dai, Juntao Tan, Yifan Zhang, Dingjie Song, Dimitris N. Metaxas, Silvio Savarese, Ran Xu, Zeyuan Chen}
}
\setcitestyle{authoryear,round,semicolon}

\renewcommand{\headrulewidth}{0.55pt}
\renewcommand{\headrule}{%
  \hbox to\headwidth{\color{SFRuleBlue}\leaders\hrule height \headrulewidth\hfill}%
}

\titleformat{\section}
  {\large\bfseries\color{SFBlue}}
  {\thesection}{0.85em}{}
\titleformat{\subsection}
  {\normalsize\bfseries\color{SFBlue}}
  {\thesubsection}{0.75em}{}
\titleformat{\subsubsection}
  {\normalsize\bfseries\color{SFBlue}}
  {\thesubsubsection}{0.65em}{}
\titlespacing*{\section}{0pt}{2.0ex plus 0.5ex minus 0.2ex}{0.8ex}
\titlespacing*{\subsection}{0pt}{1.6ex plus 0.4ex minus 0.2ex}{0.55ex}
\titlespacing*{\subsubsection}{0pt}{1.4ex plus 0.3ex minus 0.2ex}{0.45ex}

\newtcolorbox{alprompt}[1]{
    boxrule = 1pt,
    fontupper = \small\tt,
    fonttitle = \bf\color{black},
    arc = 2pt,
    rounded corners,
    colframe = black,
    colbacktitle = white!97!yellow,
    colback = white!97!yellow,
    title = #1,
}

\definecolor{citecolor}{HTML}{0064E0}
\definecolor{darkgreen}{rgb}{0.0, 0.5, 0.0}
\definecolor{darkgray}{gray}{0.4}
\definecolor{reviewer_1}{rgb}{0.5, 0.0, 0.0}
\definecolor{reviewer_2}{rgb}{0.0, 0.0, 0.5}
\definecolor{teal}{rgb}{0.0, 0.5, 0.5}

\definecolor{improvecolor}{RGB}{34,139,34} 

\definecolor{general_review}{rgb}{0.098, 0.337, 0.600}

\definecolor{reviewer_1}{RGB}{52, 152, 219}   
\definecolor{reviewer_2}{RGB}{46, 204, 113}   
\definecolor{reviewer_3}{RGB}{241, 196, 15}   
\definecolor{reviewer_4}{RGB}{149, 165, 166}

\definecolor{mylightgreen}{RGB}{144,238,144}
\definecolor{mylightblue}{RGB}{173,216,230}

\definecolor{outerboxcolor}{gray}{0.90} 
\definecolor{innerboxcolor}{rgb}{1,1,1}

\definecolor{takeawayblue}{RGB}{226,238,252}

\newtcolorbox{takeawaybox}[1][\thesubsection]{
    enhanced,
    colback=takeawayblue,
    colframe=black,
    boxrule=0.9pt,
    arc=2mm,
    left=3mm, right=3mm, top=4mm, bottom=2mm,
    before skip=10pt, after skip=8pt,
    title={Section~#1 Takeaways},
    fonttitle=\bfseries,
    coltitle=white,
    attach boxed title to top left={xshift=3mm,yshift=-2mm},
    boxed title style={
        colback=black, colframe=black,
        boxrule=0pt, arc=1mm,
        left=3mm, right=3mm, top=0.8mm, bottom=0.8mm
    }
}

\newenvironment{takeaways}[1][\thesubsection]
    {\begin{takeawaybox}[#1]
     \begin{itemize}[leftmargin=1.2em,labelsep=0.35em,
                    itemsep=1pt,topsep=0pt,parsep=0pt]}
    {\end{itemize}\end{takeawaybox}}

\definecolor{CaseTitle}{HTML}{DCE7F2}
\definecolor{CaseBody}{HTML}{F6F9FC}
\definecolor{CaseBorder}{HTML}{9AAEC3}
\definecolor{CaseInk}{HTML}{263B53}

\definecolor{NegativeTitle}{HTML}{ECE2E6}
\definecolor{NegativeBody}{HTML}{FCF8F9}
\definecolor{NegativeBorder}{HTML}{B9A4AD}
\definecolor{NegativeInk}{HTML}{634C57}

\tcbset{
    negativecase/.style={
        colbacktitle=NegativeTitle,
        colback=NegativeBody,
        colframe=NegativeBorder,
        coltitle=NegativeInk
    }
}

\newtcolorbox{operatorcase}[2][]{
    enhanced,
    breakable,
    colback=CaseBody,
    colframe=CaseBorder,
    colbacktitle=CaseTitle,
    coltitle=CaseInk,
    fonttitle=\bfseries,
    fontupper=\normalsize,
    title={#2},
    boxrule=0.6pt,
    arc=2mm,
    left=3mm,
    right=3mm,
    top=2mm,
    bottom=2mm,
    toptitle=1.5mm,
    bottomtitle=1.5mm,
    before skip=10pt,
    after skip=10pt,
    before upper={
        \setlength{\parindent}{0pt}
        \setlength{\parskip}{3pt}
    },
    #1
}

\newcommand{\CaseField}[2]{%
    \par\noindent\textbf{#1.}\enspace#2\par
}

\newcommand{\CaseStep}[2]{%
    \par\noindent
    \hangindent=3.5em
    \hangafter=1
    \makebox[3.5em][l]{\textbf{#1}}#2\par
}

\definecolor{TBHeader}{HTML}{DFEAF4}
\definecolor{ProHeader}{HTML}{E9E3F1}
\definecolor{DeepHeader}{HTML}{F3E9DA}

\definecolor{OperaFill}{HTML}{EDF6F3}
\definecolor{OperaDark}{HTML}{2F756C}

\definecolor{GainFill}{HTML}{FBF5E9}
\definecolor{GainBest}{HTML}{F1D9AE}
\definecolor{GainDark}{HTML}{9A602B}

\definecolor{OperaBest}{HTML}{287A6E}
\definecolor{OperaSecond}{HTML}{2463A6}
\definecolor{OperaBand}{HTML}{E8F1FB}
\definecolor{OperaRow}{HTML}{EAF5F2}
\definecolor{OperaGray}{HTML}{F5F5F3}

\newcommand{\opscore}[2]{#1\,{\scriptsize$\pm$\,#2}}
\newcommand{\opbest}[2]{\textcolor{OperaBest}{\textbf{#1}\,{\scriptsize$\pm$\,#2}}}
\newcommand{\opsecond}[2]{\textcolor{OperaSecond}{\underline{#1}\,{\scriptsize$\pm$\,#2}}}

\providecolor{caseproblem}{RGB}{140,80,0}
\providecolor{casecritic}{RGB}{176,40,40}
\providecolor{caseevidence}{RGB}{20,110,70}
\providecommand{\CaseProblem}[1]{\textcolor{caseproblem}{\textbf{#1}}}
\providecommand{\CaseKey}[1]{\textcolor{casecritic}{\textbf{#1}}}
\providecommand{\CaseEvidence}[1]{\textcolor{caseevidence}{\textbf{#1}}}

\newlist{contract}{itemize}{1}
\setlist[contract]{label=\textbullet, leftmargin=*, labelindent=0pt, align=left,
                   itemsep=1pt, topsep=2pt, parsep=0pt}

\newcommand{\model}{\mbox{Opera}\xspace}

\definecolor{mygray}{gray}{0.9}  

\definecolor{PromptNavy}{HTML}{153E73}
\definecolor{PromptBlue}{HTML}{286AC7}
\definecolor{PromptBlueBG}{HTML}{EDF3FC}
\definecolor{PromptGray}{HTML}{898781}
\definecolor{PromptGrayBG}{HTML}{F3F2EF}
\definecolor{PromptTeal}{HTML}{458C85}
\definecolor{PromptTealBG}{HTML}{EDF6F4}
\definecolor{PromptBorder}{HTML}{D8DADF}

\newtcblisting{promptsection}[3]{%
    enhanced,
    breakable,
    listing only,
    listing engine=listings,
    title={#1},
    fonttitle=\sffamily\bfseries\small,
    coltitle=#2,
    colbacktitle=#3,
    colback=#3,
    colframe=PromptBorder,
    boxrule=0.4pt,
    titlerule=0pt,
    borderline west={2pt}{0pt}{#2},
    arc=2pt,
    boxsep=0pt,
    left=9pt,
    right=9pt,
    top=5pt,
    bottom=7pt,
    toptitle=6pt,
    bottomtitle=2pt,
    before skip=2pt,
    after skip=2pt,
    listing options={
        basicstyle=\ttfamily\small,
        columns=fullflexible,
        keepspaces=true,
        breaklines=true,
        breakatwhitespace=true,
        showstringspaces=false,
        aboveskip=0pt,
        belowskip=0pt
    }
}

\newtcolorbox{salesforcetitlebox}{
    enhanced,
    colback=SFBoxBG,
    colframe=SFBoxBorder,
    boxrule=0.65pt,
    arc=4mm,
    left=6mm,
    right=6mm,
    top=4.5mm,
    bottom=4.5mm,
    before skip=2pt,
    after skip=15pt
}

\begin{document}
\thispagestyle{fancy}

\begin{salesforcetitlebox}
{\raggedright\fontsize{19.5}{22.5}\selectfont\bfseries\color{SFBlue}
Opera: A Verbal Critic Framework for Long-horizon Coding Agents\par}

\vspace{0.65em}

{\raggedright\normalsize\bfseries
Kai Mei$^{1,2}$,
Zhiyuan Hu$^{1}$,
Yutong Dai$^{1}$,
Juntao Tan$^{1}$,
Yifan Zhang$^{1}$,
Dingjie Song$^{1,3}$,
Dimitris N. Metaxas$^{2}$,
Silvio Savarese$^{1}$,
Ran Xu$^{1}$,
Zeyuan Chen$^{1}$\par}

\vspace{0.55em}

{\raggedright\normalsize\bfseries
$^{1}$Salesforce AI Research,
$^{2}$Rutgers University,
$^{3}$Lehigh University \par}

\vspace{1.05em}

{\large\bfseries\color{SFBlue}Abstract\par}
\vspace{0.25em}
Long-horizon coding agents need timely corrections, yet feedback can be ineffective or even harmful when it misjudges ongoing work or fails to address the underlying problem. Existing critics focus on evaluating trajectories and generating feedback, but rarely track what happens after feedback is delivered. We present \model{}, a verbal critic framework that treats each correction as a persistent note, followed until the diagnosed problem is resolved. \model{} decides when to review through periodic and event-driven triggers, diagnoses issues with typed operators, audits feedback against visible evidence before delivery, and tracks the agent's subsequent actions to distinguish mere compliance from actual resolution. As a test-time critic, \model{} improves the resolve rate of non-critic agents by up to 12.4, 15.0, and 8.9 percentage points on Terminal-Bench 2.1, a SWE-Bench Pro subset, and DeepSWE v1.1, respectively, across four policy models, and achieves the highest mean resolve rate among competitive critic baselines on all three benchmarks, and also improves policy models when the policy critiques itself. Beyond inference, \model{}-guided rollouts provide approximately on-policy training data: fine-tuning Qwen3.5-9B on them improves its resolve rate on held-out SWE-Bench Pro repositories by 10.2 percentage points without a critic at inference time, matching fine-tuning on rollouts from a stronger model, while preserving its performance when switching harness, i.e., from Openhands to Terminus-2, which the latter substantially degrades. 
Our code is available at: https://github.com/dongyuanjushi/Opera.
\end{salesforcetitlebox}

\section{Introduction}
\label{sec:intro}

Coding agents tackle repository-level tasks through code exploration, editing, and testing~\citep{yang2024sweagent,wang2024openhands}. Over long horizons, they often repeat unsuccessful actions or explore redundantly~\citep{gandhi2025sweprm}, and may report completion before the task is actually verified~\citep{tang2026misalignment}. A critic that diagnoses such problems during execution is a natural remedy, but it must judge unfinished work, where incomplete evidence blurs the line between an actual mistake and a reasonable intermediate step.

Prior work provides strong foundations for feedback and reflection, including Self-Refine~\citep{madaan2023selfrefine}, Reflexion~\citep{shinn2023reflexion}, and CRITIC~\citep{gou2023critic}, and for coding agents, SWE-PRM~\citep{gandhi2025sweprm}, SWE-Search~\citep{antoniades2024swesearch}, and Agentic Rubrics~\citep{raghavendra2026agenticrubrics}. We move further to view critic feedback as an \emph{intervention} whose value is revealed only afterwards, which raises three challenges: \ding{172} \textbf{\emph{When should the critic intervene?}} Periodic feedback may arrive too late, while not every execution event requires correction. \ding{173} \textbf{\emph{Is the feedback justified?}} A plausible diagnosis can misread partial evidence and derail work that would otherwise succeed~\citep{vasudev2026intervention}, and LLM judges remain noisy even when verifying well-specified criteria~\citep{peng2026ruverbench}.  \ding{174} \textbf{\emph{Did the intervention work?}} An agent may follow a suggestion yet leave the underlying failure intact.

To address the above challenges, we introduce \model{}, a verbal critic framework that manages each correction as a persistent note: opened when an issue is found, delivered only when justified, tracked as execution proceeds, and closed once evidence supports its resolution. \model{} combines periodic and event-driven triggers with the option to stay silent (\ding{172}); pairs an operator critic, which ties each issue to evidence, a corrective direction, and a resolution criterion, with an audit that screens feedback before delivery (\ding{173}); and maintains a finding note that tracks adherence separately from resolution (\ding{174}). \model{} interacts with agents only through execution traces and natural-language feedback, so it plugs into existing harnesses without updating agent weights.

At test time, \model{} acts as a teacher that guides the agent's self-reflection. Across Terminal-Bench 2.1, a 100-task SWE-Bench Pro subset, and DeepSWE v1.1, \model outperforms four competitive critic baselines and shows consistent improvement on four policy models with different harnesses. Beyond inference, the same process yields training data that distillation from stronger models lacks: trajectories consisting of the student's own actions, together with targeted diagnoses of its errors. 

Our contributions are threefold:
\begin{itemize}[leftmargin=1.2em,labelsep=0.35em,
                    itemsep=1pt,topsep=0pt,parsep=0pt]
    \item \textbf{Framework.} We formulate verbal criticism as managing persistent corrective notes and introduce \model{}, a critic framework that decides when to review, audits feedback before delivery, and follows each correction until it is resolved during long-horizon runs of coding agents.
    \item \textbf{Test-time critic.} \model{} improves task resolve rate by up to 12.4, 15.0, and 8.9 pp on Terminal-Bench 2.1, SWE-Bench Pro, and DeepSWE v1.1 across four policy models, and outperforms competitive critic baselines. It improves most policy models with both weak and strong critic models, including the policy model critiquing itself.
    \item \textbf{Training recipe.} \model{}-guided student rollouts provide approximately on-policy training data. Fine-tuning Qwen3.5-9B on them matches distillation from a stronger model on held-out SWE-Bench Pro repositories (+10.2 pp) while preserving its Terminal-Bench 2.1 performance.
\end{itemize}

\begin{figure}
    \centering
    \includegraphics[width=\linewidth]{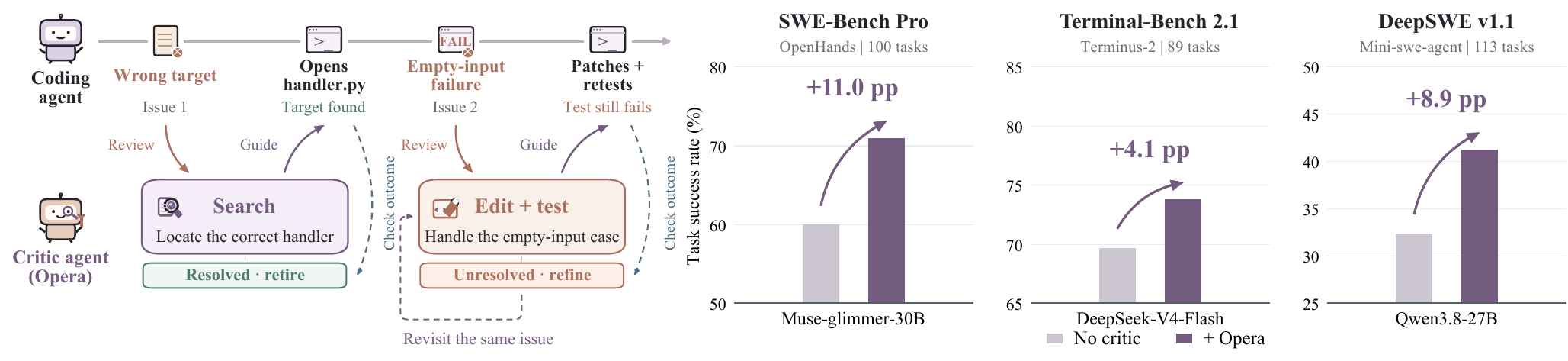}
    \caption{Illustration of \model{}. \textbf{Left:} \model{} discovers and tracks issues at different stages (search, edit, test, etc.) until it is resolved or refined in the long-horizon turns. \textbf{Right:} \model{} improves task resolve rate across benchmarks, harnesses, and policy models.}
    \label{fig:intro_motivation}
    \vspace{-10pt}
\end{figure}

\section{Related Work}
\label{sec:related_work}

\subsection{Critics and Verifiers for Coding Agents}
\label{sec:related_work_critics}

Early approaches improve LLM outputs through iterative self-feedback \citep{madaan2023selfrefine}, verbal reflection across attempts \citep{shinn2023reflexion}, and tool-grounded critique \citep{gou2023critic}. For repository-level coding, SWE-PRM~\citep{gandhi2025sweprm} periodically reviews agent trajectories and provides taxonomy-guided corrective feedback. Agentic Rubrics~\citep{raghavendra2026agenticrubrics} constructs repository-specific criteria for evaluating candidate patches, while SWE-Shepherd~\citep{dihan2026sweshepherd} uses a trained process reward model to guide intermediate action selection. However, RuVerBench~\citep{peng2026ruverbench} identifies substantial noise in LLM-based rubric verification, motivating explicit quality control over critic feedback.

\subsection{Test-Time Scaling for Coding Agents}
\label{sec:related_work_test_time_scaling}
Test-time scaling allocates additional inference computation to exploration, evaluation, and refinement. Tree of Thoughts~\citep{yao2023treeofthoughts} explores alternative reasoning states, while Language Agent Tree Search~\citep{zhou2023lats} combines tree search with reflection and environment feedback. For coding, SWE-Search~\citep{antoniades2024swesearch} applies Monte Carlo tree search to repository-level tasks, and
S$^{*}$~\citep{li2025sstar} combines parallel sampling, sequential refinement, and execution-grounded selection.
Recent work uses rollout summaries for trajectory selection and reuse~\citep{kim2026scalingagenticcoding}, differential testing for candidate selection~\citep{he2026diffcodegen}, and an LLM orchestrator for adaptive solver allocation and answer
synthesis~\citep{qin2026atlas}. \model allocates additional inference computation to diagnosing and correcting an ongoing coding trajectory. Its hybrid review schedule and audited interventions connect compute allocation to when feedback is warranted and how it should guide subsequent execution. 

\section{\model: Verbal Critic Framework through Persistent Notes}
\label{sec:method}

\begin{figure}[t]
    \centering
    \includegraphics[width=0.95\linewidth]{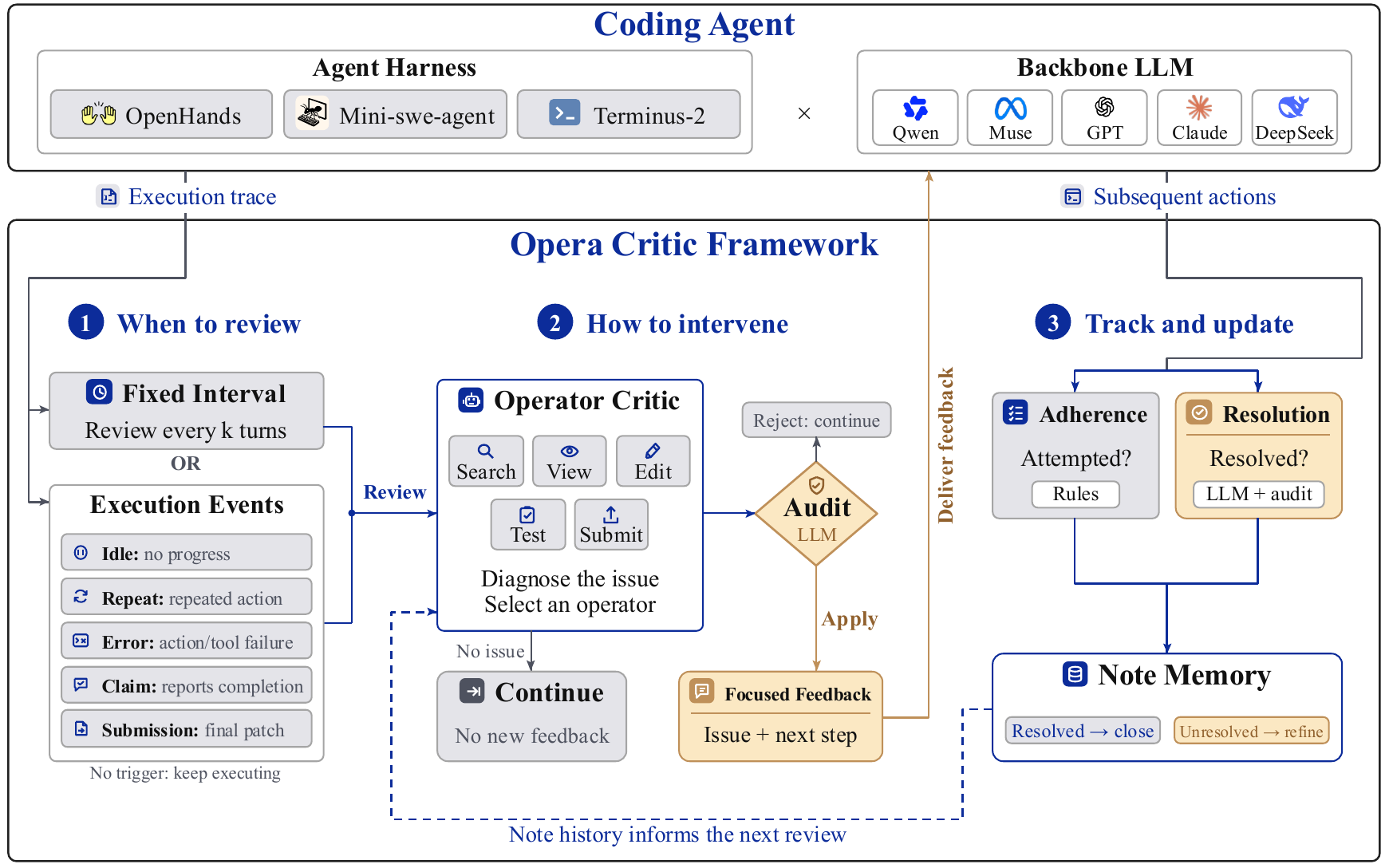}
    \caption{Overview of \model{}. \ding{182} Reviews are triggered at fixed intervals or by execution events. \ding{183} The operator critic diagnoses one issue with a typed operator, and an audit decides whether the feedback is delivered. \ding{184} Adherence and resolution are tracked separately; resolved notes are closed, unresolved notes are refined, and note history conditions the next review.}
    \vspace{-10pt}
    \label{fig:overview}
\end{figure}

An effective critic must decide when to review, deliver feedback the agent can act on, and verify afterwards that the problem is resolved. \model{} organizes supervision around a persistent \emph{note}: one diagnosed issue, its current guidance, and a fixed resolution criterion. Persistence lets the critic follow up on its own verbal feedback across reviews; the fixed criterion ensures that resolution is judged against the original problem rather than the latest guidance. \autoref{fig:overview} shows how \model{} addresses the three challenges from Section~\ref{sec:intro}: when to review (\S\ref{sec:method_scheduling}), how to intervene (\S\ref{sec:method_operators}), and how to track and update notes (\S\ref{sec:method_memory}).

\subsection{Hybrid Review Scheduling}
\label{sec:method_scheduling}

Fixed-interval review reacts slowly: an agent may repeat a failing command several times before the next review. Event-triggered review reacts quickly but misses silent failures, such as confidently editing the wrong file. \model{} combines both. Unlike SWE-PRM~\citep{gandhi2025sweprm}, which reviews at fixed intervals, \model{} performs a periodic review every $k$ turns plus immediate reviews on five events: \emph{Idle} (no visible progress), \emph{Repeat} (repeated actions), \emph{Error} (failed actions or tool calls), \emph{Claim} (reported completion), and \emph{Submission} (final patch). Simultaneous signals produce a single review, and an event-triggered review resets the periodic timer. A review does not imply an intervention: the critic may let the agent continue without feedback.

\subsection{Operator-Typed Review}
\label{sec:method_operators}

\begin{figure}[t]
    \centering
    \includegraphics[width=0.95\linewidth]{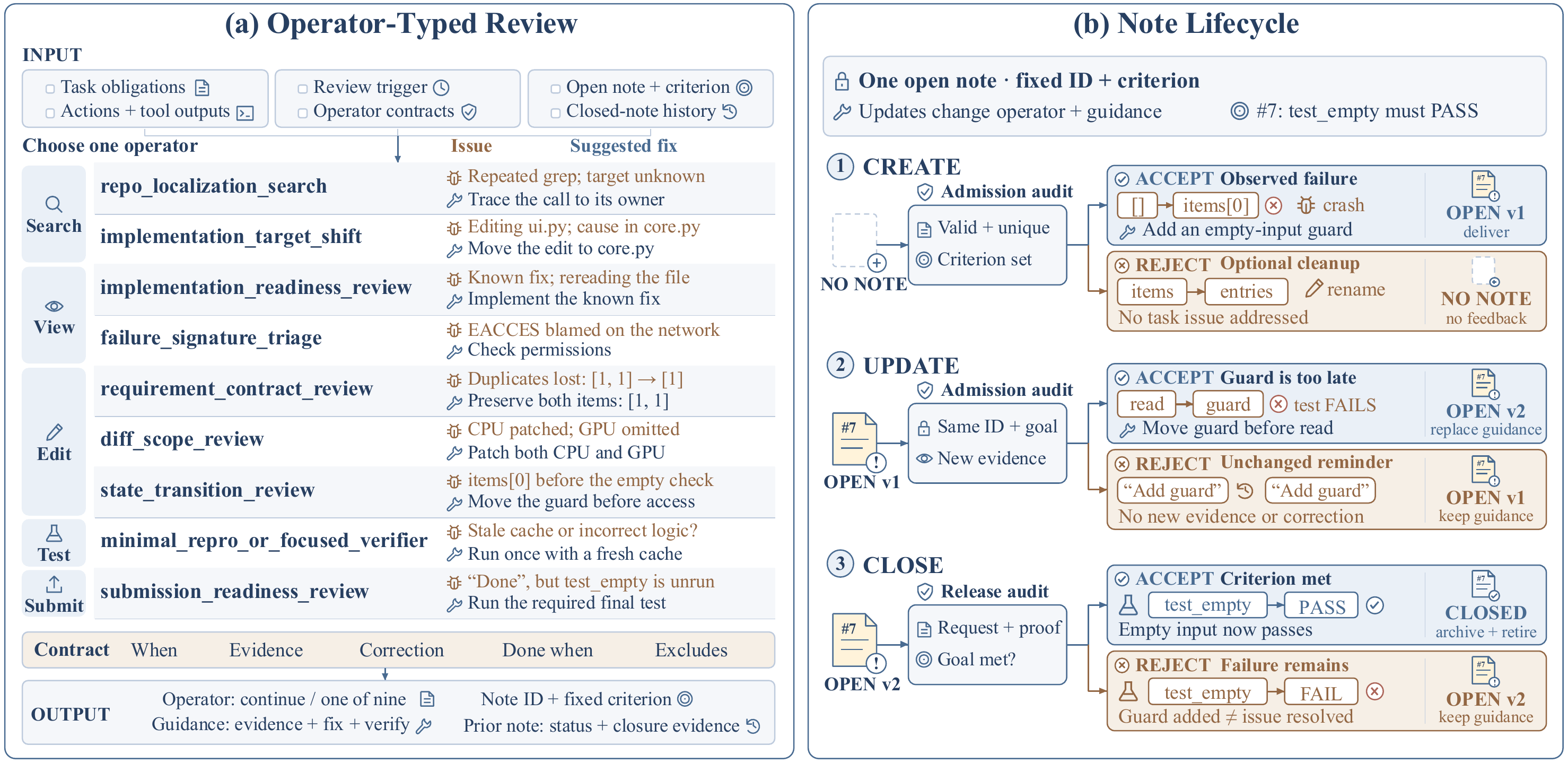}
    \caption{Core mechanisms of \model{}. (a) Nine operators, grouped by repair stage and specified by contracts, map an observed issue to a targeted fix. (b) Admission audits gate note creation and updates, rejecting unsupported or redundant feedback; the release audit closes a note only when its fixed criterion is met.}
    \label{fig:state_transition}
    \vspace{-15pt}
\end{figure}

Free-form critique is often generic (e.g., ``consider edge cases'') or mixes several concerns, leaving the agent little to act on. \model{} instead requires each diagnosis to use a typed \emph{operator}. The operators follow the stages of code repair, namely localization, inspection, editing, and verification~\citep{xia2024agentless,bouzenia2024repairagent}, and cover common failure types such as misunderstood requirements, incorrect edit scope, and faulty logic~\citep{tang2026misalignment}. Nine operators span five stages (\autoref{fig:state_transition}a):

\begin{itemize}[leftmargin=*, labelsep=0.4em, itemsep=2pt, topsep=2pt, parsep=0pt]
    \item \textbf{Search}: \textit{repo\_\allowbreak localization\_\allowbreak search} (target not yet identified); \textit{implementation\_\allowbreak target\_\allowbreak shift} (editing the wrong target).
    \item \textbf{View}: \textit{implementation\_\allowbreak readiness\_\allowbreak review} (re-inspecting code when the fix is already clear); \textit{failure\_\allowbreak signature\_\allowbreak triage} (misread tool output).
    \item \textbf{Edit}: \textit{requirement\_\allowbreak contract\_\allowbreak review} (requirement violated); \textit{diff\_\allowbreak scope\_\allowbreak review} (missing or unrelated changes); \textit{state\_\allowbreak transition\_\allowbreak review} (faulty execution logic).
    \item \textbf{Test}: \textit{minimal\_\allowbreak repro\_\allowbreak or\_\allowbreak focused\_\allowbreak verifier} (missing evidence needed for the next decision). 
    \item \textbf{Submit}: \textit{submission\_\allowbreak readiness\_\allowbreak review} (completion claimed without sufficient evidence).
\end{itemize}

All operators remain available throughout execution, since agents often revisit earlier stages. Following Design by Contract~\citep{meyer1992contract}, each operator is specified by a contract stating when it applies, the evidence it requires, the correction it may request, when the issue counts as resolved, and which problems it excludes. The exclusion clause keeps operators from overlapping, and the resolution clause supplies the criterion against which the note is later checked.

\textbf{Selection and output.}
Given the task, execution history, review trigger, and current note, the critic either returns \textit{continue} when the agent is making useful progress, since unnecessary interventions can disrupt trajectories that would otherwise succeed~\citep{vasudev2026intervention}, or selects one supported issue and the smallest correction for it. Restricting each review to one issue avoids competing instructions and keeps later outcomes attributable to a specific correction. Each proposal names an operator, cites evidence, and specifies the location, the current behavior, the required change, and how to verify it. Independently, the critic reports whether the open note remains unresolved, so follow-up continues even without new guidance.

\subsection{Audited Note Management}
\label{sec:method_memory}

Inspired by reflection memory and persistent goals~\citep{shinn2023reflexion,cohen1990commitment}, each note is either \textit{open} or \textit{closed}, with at most one open note at a time (\autoref{fig:state_transition}b). This keeps the agent focused on a single correction and makes outcomes attributable. Because the critic is itself an LLM whose diagnoses may be plausible but wrong, two audits govern the note lifecycle. Both only accept or reject proposals without rewriting them, acting as verifiers rather than co-authors.

\textbf{Opening and updating.}
The critic may open a note when none is open, or update the open note when new evidence changes the diagnosis or suggests a clearer next step. Updates may change the operator and guidance but preserve the note's identity and criterion, so the goal stays fixed while the advice improves. After validity and duplicate checks, the audit verifies that the evidence supports the diagnosis, the correction is necessary, and the proposed check can establish resolution~\citep{vasudev2026intervention}. It rejects, for example, optional cleanups unrelated to the task and reminders that repeat guidance without new evidence. Rejected proposals leave the current note unchanged.

\textbf{Tracking and closing.}
\model{} tracks two distinct questions after delivery. \emph{Adherence}, checked by rules, asks whether the agent attempted the correction. \emph{Resolution}, judged against the fixed criterion, asks whether the original problem is gone. They are separated because adherence does not imply resolution: in \autoref{fig:state_transition}b, the agent adds the suggested guard, yet the required test still fails. Closing a note requires a \emph{release audit} with supporting evidence, namely execution results for behavioral requirements or visible edits for static ones. Accepted closure archives the note and removes its guidance from the agent's context. Closure is processed before admission, so one review can close an issue and open the next.

\section{Evaluation}
\subsection{Benchmarks and Setup}
\label{sec:setup}
We evaluate on Terminal-Bench 2.1 (89 tasks)~\citep{terminalbench2026v21} using Terminus-2, SWE-Bench Pro (100 tasks)~\citep{deng2025swebenchpro} using OpenHands~\citep{wang2024openhands}, and DeepSWE v1.1 (113 tasks)~\citep{huang2026deepswe} using mini-swe-agent~\cite{yang2024sweagent}. For SWE-Bench Pro, we apply task corrections informed by OpenAI's benchmark audit~\citep{openai2026codingsignals} and sample 100 tasks from the corrected set and cover all of the repos included in the raw swebench-pro. Across all three benchmarks, we employ Harbor~\citep{harbor2026} to manages tasks, and Pier~\citep{datacurve2026pier} to provides network isolation for anti-cheating. For policy model, we use Qwen3.5-9B (thinking) \citep{qwen2026qwen35}, Muse-glimmer-30B (high) \citep{meta2026museglimmer}, Qwen3.8-27B (xhigh) \citep{qwen2026qwen38} and Deepseek-V4-Flash-0731 (max) \citep{deepseek2026v4} for our evaluations. We use GPT-5.6-Sol (high) \citep{openai2026gpt56sol} as the default critic model for all of our experiments if not specifically mentioned, results of using other models are also reported in Section \ref{sec:sensitivity_to_critic}. For timeouts, we employ the task-specific timeouts configured in Harbor's official dataset cards \footnote{https://hub.harborframework.com/datasets}. 

\subsection{Overall Task Performance}

\begin{wraptable}{r}{0.53\textwidth}
    \vspace{-25pt}
    \centering
    \footnotesize
    \setlength{\tabcolsep}{3pt}
    \renewcommand{\arraystretch}{1.1}
    \caption{Comparison with critic baselines (policy: Qwen3.8-27B). Resolve rate (\%, mean $\pm$ std) on Terminal-Bench 2.1 (TB-2.1, 89 tasks), a 100-task subset of SWE-Bench Pro (SWE-Pro), and DeepSWE v1.1 (113 tasks). Green bold / blue underline: best / second-best. }
    \label{tab:critic_comparison}
    \begin{tabular}{@{}lccc@{}}
        \toprule
        Method & TB-2.1 & SWE-Pro & DeepSWE \\
        \midrule
        Non-critic      & \opscore{65.9}{1.7} & \opscore{75.3}{1.2} & \opscore{32.4}{1.0} \\
        \midrule
        SWE-PRM         & \opscore{68.9}{3.2} & \opscore{77.7}{2.5} & \opscore{36.6}{3.1} \\
        SWE-Search      & \opscore{70.0}{4.7} & \opscore{77.3}{3.8} & \opscore{38.6}{4.4} \\
        LLM-as-verifier & \opscore{71.2}{2.6} & \opsecond{78.7}{2.3} & \opscore{38.9}{2.3} \\
        Agentic Rubrics & \opsecond{71.9}{4.5} & \opscore{78.3}{3.2} & \opsecond{39.5}{2.8} \\
        \midrule
        \rowcolor{OperaRow}
        \textbf{\model{} (ours)} & \opbest{73.8}{2.3} & \opbest{79.3}{3.5} & \opbest{41.3}{3.1} \\
        Gain (pp)       & $+7.9$ & $+4.0$ & $+8.9$ \\
        \bottomrule
    \end{tabular}
\end{wraptable}

\textbf{Comparison with critic baselines.} We compare \model{} with the non-critic agent and four representative critics, all using Qwen3.8-27B as the policy model (\autoref{tab:critic_comparison}). They fall into two groups by what they judge and what signal they return. \emph{Process critics} evaluate the ongoing trajectory: SWE-PRM~\citep{gandhi2025sweprm} periodically reviews recent steps and returns taxonomy-guided verbal feedback, and SWE-Search~\citep{antoniades2024swesearch} scores the last action with a numerical value and a natural-language assessment, originally used to guide tree search. \emph{Outcome verifiers} score the agent's solution: LLM-as-verifier~\citep{kwok2026llm} assigns fine-grained scalar scores, and Agentic Rubrics~\citep{raghavendra2026agenticrubrics} scores patches against a repository-grounded rubric checklist without executing tests. For \emph{Process Critics}, we adapt them to use the same critic schedule as \model{} to ensure a fair comparison, more implementation details are included in Appendix \ref{app:impl_details}. \autoref{tab:critic_comparison} shows that \model{} achieves the highest mean resolve rate on all three benchmarks. Its margin over the strongest baseline is 1.9 pp on Terminal-Bench 2.1 and 1.8 pp on DeepSWE, but only 0.6 pp on SWE-Bench Pro, where the non-critic agent is already strong and all critics yield similar, modest gains; This improvement suggests the the advantages of \model{}'s design on the two aspects: it intervenes during execution like a process critic, while each note carries an explicit resolution criterion like the output verifier before closure.

\begin{figure}[tb!]
    \centering
    \includegraphics[width=0.95\linewidth]{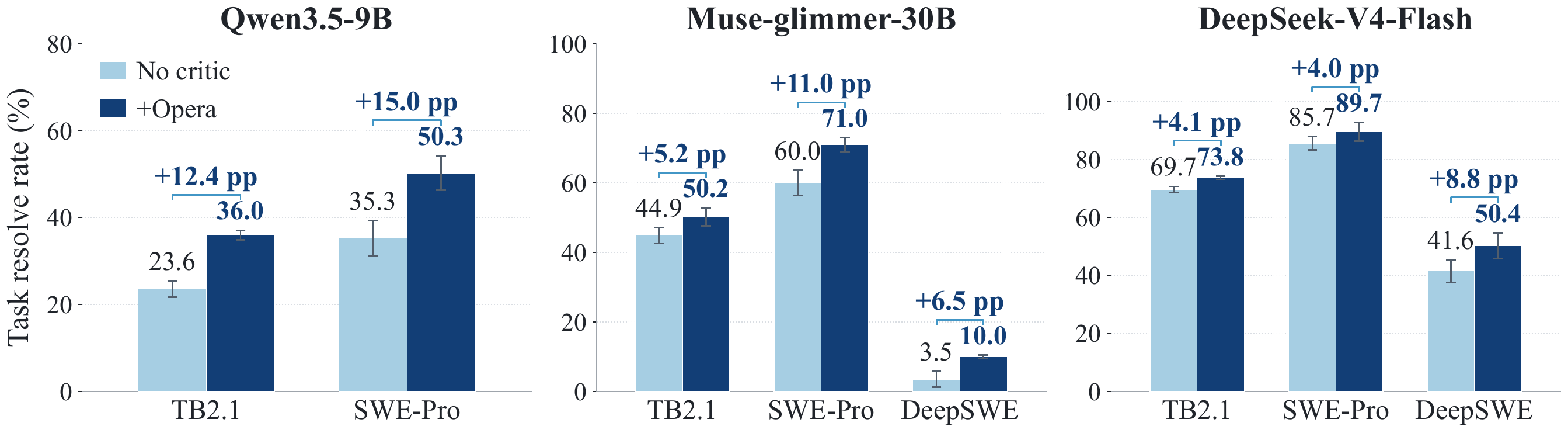}
    \caption{Task resolve rate of \model{} across policy models (critic: GPT-5.6-Sol). Gains (pp) are generally larger for weaker backbones and on benchmarks with more headroom. Qwen3.5-9B resolves no DeepSWE task with or without the critic. Error bars denote sample standard deviation.}
    \label{fig:other_backbone}
    \vspace{-2pt}
\end{figure}

\textbf{Performance on different policy models.}
\autoref{fig:other_backbone} shows that \model{} improves every policy model on every benchmark, with one exception: Qwen3.5-9B resolves no DeepSWE task, with or without the critic. The gains are shaped by how much room the agent leaves for correction. Weaker agents generally benefit more: Qwen3.5-9B gains 12.4 and 15.0 pp on Terminal-Bench 2.1 and SWE-Bench Pro, and gains tend to shrink as the non-critic resolve rate rises. Headroom matters more than model strength alone: DeepSeek-V4-Flash gains little on benchmarks where it is already strong, but substantially on the harder DeepSWE. Yet a critic cannot substitute for missing capability. DeepSWE consists of original, long-horizon engineering tasks, and Qwen3.5-9B resolves no DeepSWE task in the evaluated runs, with or without the tested critics. Muse Glimmer-30B on DeepSWE shows a milder version of this effect: starting from a low base, feedback nearly triples its resolve rate, but the absolute gain remains limited. A critic is therefore most valuable when the agent is capable enough to act on feedback but still makes errors it cannot correct on its own.

\subsection{Recipes for Training Open-weight Coding Agents}
\label{sec:training}

Beyond inference-time assistance, \model{} can generate training data in which the agent receives targeted diagnoses and takes corrective actions. We ask whether such data can be used to train Qwen3.5-9B to learn from error correction without a critic at inference time.

\textbf{Setup.} We split SWE-Bench Pro by repository into 270 training tasks and 108 held-out tasks (\autoref{tab:sft_data_distribution}), and additionally evaluate on Terminal-Bench 2.1, which differs from the training data in task format and harness. We compare two trajectory sources: \emph{\model{}-guided student rollouts}, with Qwen3.5-9B as the policy and Qwen3.8-27B as the critic, and \emph{teacher rollouts}, with Qwen3.8-27B as the policy. The same model thus supplies the expertise in both settings, either by critiquing or by acting. We select 67 training tasks on which the base Qwen3.5-9B rarely succeeds but both sources contain at least one successful trajectory, so that both sources supervise the same tasks (selection criteria and more training details are in Appendix~\ref{app:training}). 

\begin{table*}[tb!]
    \centering
    \small
    \begin{minipage}[t]{0.52\textwidth}
        \centering
        \caption{Task distribution of the SFT training split and the held-out OOD evaluation split of SWE-Bench Pro. TS/JS: TypeScript/JavaScript.}
        \label{tab:sft_data_distribution}
        \setlength{\tabcolsep}{3pt}
        \renewcommand{\arraystretch}{1.16}
        \begin{tabularx}{\linewidth}{@{}l>{\raggedright\arraybackslash}Xrrrrr@{}}
            \toprule
            Split & Repositories & Py & Go & TS & JS & Total \\
            \midrule
            Train & qutebrowser, ansible, teleport, flipt, vuls, webclients, NodeBB & 99 & 115 & 32 & 24 & 270 \\
            \addlinespace[2pt]
            Held-out & openlibrary, navidrome, element-web, tutanota & 47 & 30 & 31 & 0 & 108 \\
            \bottomrule
        \end{tabularx}
    \end{minipage}
    \hfill
    \begin{minipage}[t]{0.47\textwidth}
        \centering
        \caption{Resolve rate (\%) of Qwen3.5-9B trained on different trajectory sources, evaluated without a critic on held-out SWE-Bench Pro tasks and Terminal-Bench 2.1. Mean $\pm$ std over three runs.}
        \label{tab:training}
        \setlength{\tabcolsep}{4pt}
        \renewcommand{\arraystretch}{1.16}
        \begin{tabularx}{\linewidth}{@{}>{\raggedright\arraybackslash}Xcc@{}}
            \toprule
            Training data & SWE-Pro & TB2.1 \\
            \midrule
            None (base)            & $28.7 \pm 1.9$ & $23.6 \pm 1.9$ \\
            Qwen3.8-27B rollouts   & $\mathbf{39.5 \pm 1.4}$ & $7.9 \pm 1.9$ \\
            \model{}-guided (ours) & $38.9 \pm 2.4$ & $\mathbf{25.8 \pm 2.2}$ \\
            \bottomrule
        \end{tabularx}
    \end{minipage}
\end{table*}

\textbf{Results and analysis.} Both sources improve the held-out SWE-Bench Pro resolve rate by about 10 pp, with no meaningful difference between them (\autoref{tab:training}). On Terminal-Bench 2.1, however, fine-tuning on teacher rollouts reduces the resolve rate from 23.6\% to 7.9\%, whereas \model{}-guided rollouts preserve it (25.8\%). Both sources thus improve the target task similarly, but only the \model{}-guided data does so without degrading performance on a benchmark with a different task format and harness. We attribute this mainly to whose actions the data contain. Teacher rollouts are off-policy for the student, and fitting them shifts the student toward the teacher's distribution; forgetting is known to grow with such shift~\citep{shenfeld2025rlrazor}, and fine-tuning on self-generated data mitigates it~\citep{yang2024sdft}. \model{}-guided rollouts are approximately on-policy: every action comes from the student, and the critic's influence enters only through notes in its context. Such data has been shown to underlie RL's robustness to forgetting~\citep{chen2025retaining}. The recipe also follows the principle of DAgger~\citep{ross2011dagger} and on-policy distillation~\citep{agarwal2024gkd}: the student receives expert corrections at the states it actually visits, which teacher trajectories rarely contain, since the teacher seldom makes the student's mistakes.

\begin{takeaways}
    \item A critic can turn a stronger model's expertise into approximately on-policy self-reflection data for a weaker student.
    \item Fine-tuning on critic-guided student trajectories matches distillation from the stronger model on held-out repositories, while preserving generalization capabilities better.
\end{takeaways}

\section{Analysis}

\subsection{Critic Behavior and Utility}

\begin{figure}[tb!]
    \centering
    \includegraphics[width=0.95\linewidth]{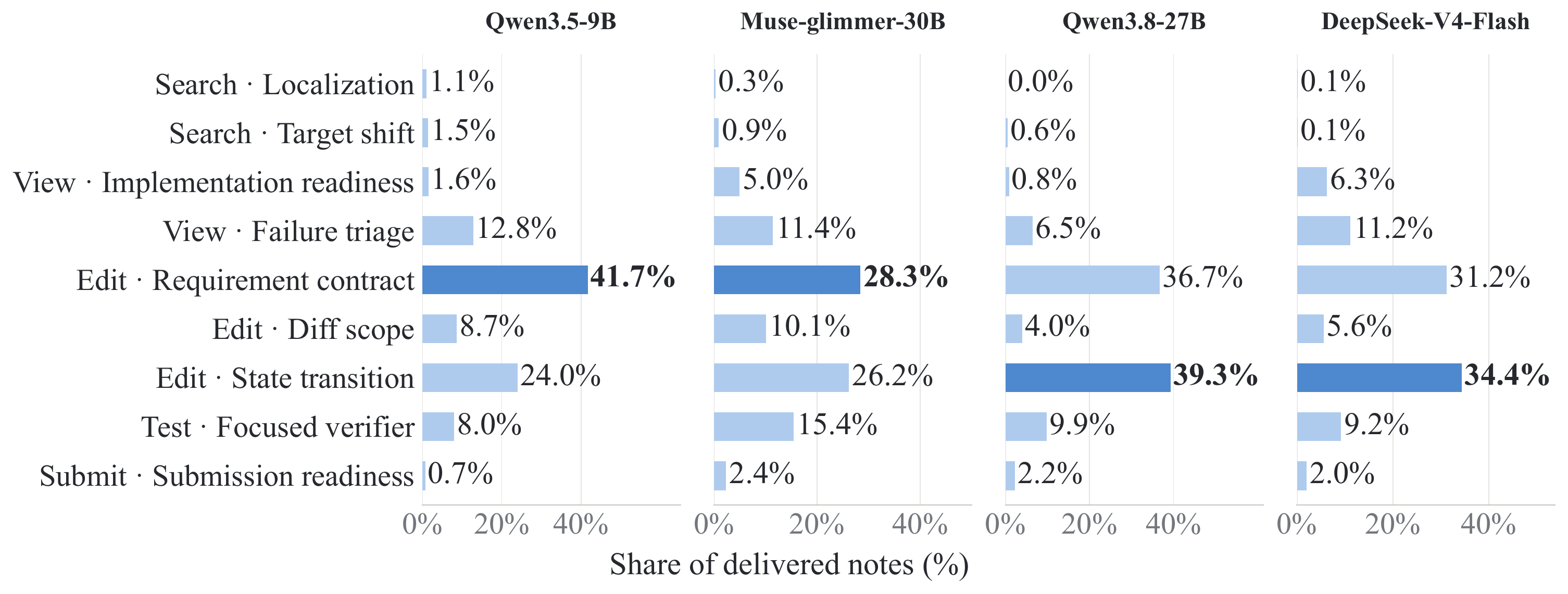}
    \caption{Distribution of admitted operators. Edit-stage operators dominate while the leading one shifts from requirement contract to state transition as model capability increases.}
    \label{fig:operator_distribution}
\end{figure}

\begin{figure}[tb!]
    \centering
    \includegraphics[width=0.9\linewidth]{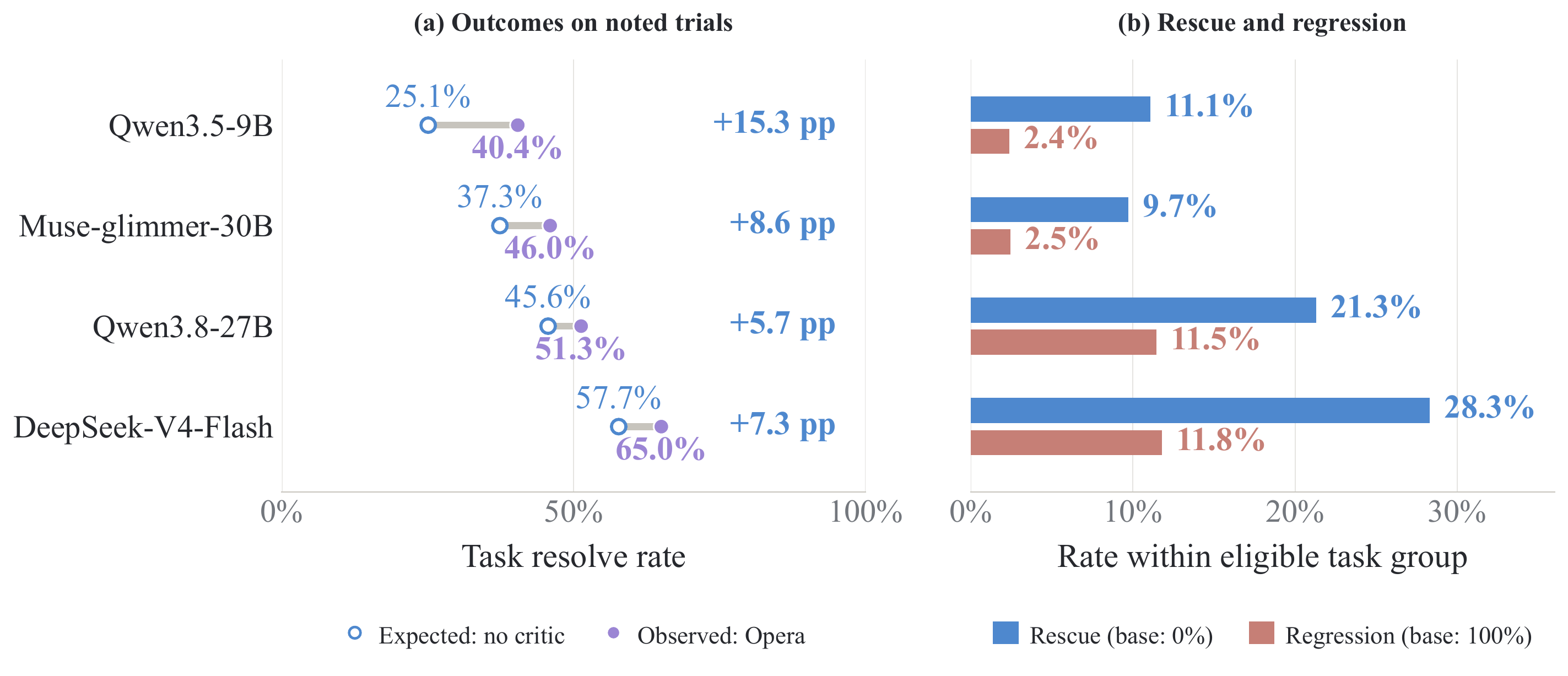}
    \caption{Task outcomes on trials receiving notes. (a) Observed \model{} success versus expected no-critic success on the same tasks. (b) Rescue rate among tasks never solved without the critic and regression rate among tasks always solved without it; the two rates use different denominators.}
    \label{fig:critic_task_outcomes}
    \vspace{-15pt}
\end{figure}

We analyze which operators the critic admits for each policy model, how agents respond to them, and how these responses affect task outcomes. Operator frequencies reflect the critic's admitted diagnoses rather than all agent errors.

\textbf{The critic mostly diagnoses implementation errors, and the dominant type is model-specific.}
Across all four models, Edit-stage operators account for most delivered notes, while Search and Submit operators are rare (\autoref{fig:operator_distribution}). This suggests that, in the trajectories the critic flags. Qwen3.5-9B mostly receives requirement-contract notes, suggesting that it often misreads what the task requires. The stronger Qwen3.8-27B and DeepSeek-V4-Flash shift toward state-transition notes, suggesting that they understand the goal but get the execution logic wrong. Muse Glimmer-30B shows a flatter distribution with more focused-verifier notes, pointing to weaker verification habits.

\textbf{Stronger models gain more rescues but also more regressions.}
On trials receiving notes, \model{} improves over the expected no-critic success for all models (\autoref{fig:critic_task_outcomes}a). The rescue--regression breakdown, however, reveals a trade-off (\autoref{fig:critic_task_outcomes}b), consistent with the recovery--disruption trade-off reported by \citet{vasudev2026intervention}. Stronger models convert feedback into more rescues of previously unsolved tasks, but also regress more often on tasks they would otherwise solve, while weaker models rarely regress but are rescued less often. One possible explanation is that stronger models act on feedback more readily (\autoref{fig:operator_response}), which amplifies both correct and unnecessary interventions; for these models, the cost of an unneeded correction is therefore higher.

\begin{takeaways}
    \item The critic mostly flags how agents change code rather than where they look.
    \item As policy capability increases, the most frequent diagnosis shifts from misread requirements to faulty execution logic.
    \item A critic's benefit is a rescue--regression trade-off, and the stronger policy models we tested rank high on both.
\end{takeaways}

\subsection{Operator ablation}

\begin{wraptable}{r}{0.52\textwidth}
    \centering
    \small
    \setlength{\tabcolsep}{3pt}
    \renewcommand{\arraystretch}{1.12}
    \vspace{-15pt}
    \caption{Operator ablation with Qwen3.8-27B. Resolve rate (\%), mean $\pm$ std over three runs; parentheses show gains over non-critic (pp).}
    \label{tab:operator_ablation}
    \resizebox{\linewidth}{!}{%
    \begin{tabular}{@{}lccc@{}}
        \toprule
        Variant & TB2.1 & SWE-Pro & DeepSWE \\
        \midrule
        Non-critic
        & $65.9 \pm 1.7$
        & $75.3 \pm 1.2$
        & $32.4 \pm 1.0$ \\
        \midrule
        \model{} (full)
        & $\mathbf{73.8 \pm 2.3}$ {\scriptsize(+7.9)}
        & $\mathbf{79.3 \pm 3.5}$ {\scriptsize(+4.0)}
        & $\mathbf{41.3 \pm 3.1}$ {\scriptsize(+8.9)} \\
        Edit-only
        & $71.9 \pm 1.6$ {\scriptsize(+6.0)}
        & $79.0 \pm 3.5$ {\scriptsize(+3.7)}
        & $40.1 \pm 1.8$ {\scriptsize(+7.7)} \\
        Non-edit
        & $70.2 \pm 0.8$ {\scriptsize(+4.3)}
        & $78.3 \pm 0.6$ {\scriptsize(+3.0)}
        & $39.5 \pm 2.7$ {\scriptsize(+7.1)} \\
        \bottomrule
    \end{tabular}%
    }
\end{wraptable}

To further examine the role of operators, we note that Edit-stage operators account for most delivered notes, e.g., about 80\% for Qwen3.8-27B (\autoref{fig:operator_distribution}). We compare two restricted variants that keep all other components of \model{} unchanged: \emph{Edit-only}, which retains only the Edit-stage operators, and \emph{Non-edit}, which retains only the Search, View, Test, and Submit operators (\autoref{tab:operator_ablation}). The full operator set achieves the highest mean on all three benchmarks, although its margin over Edit-only is small (e.g., 0.3 pp on SWE-Bench Pro) and within run-to-run variance. Both restricted variants retain most of the full gain. Notably, Non-edit operators account for only about one fifth of delivered notes but recover over half of the full gain on Terminal-Bench 2.1, three quarters on SWE-Bench Pro, and about four fifths on DeepSWE. The gains of the two variants are also not additive, as each alone recovers much of the full gain. One possible explanation is that the critic can often address the same underlying problem through different operators, so that the review and follow-up process compensates for a restricted operator set.

\begin{takeaways}
    \item Both restricted operator sets retain substantial gains, while the full set achieves the best.
\end{takeaways}

\subsection{Sensitivity to Critic Model Choice}
\label{sec:sensitivity_to_critic}

\begin{figure}[tb!]
    \centering
    \includegraphics[width=\linewidth]{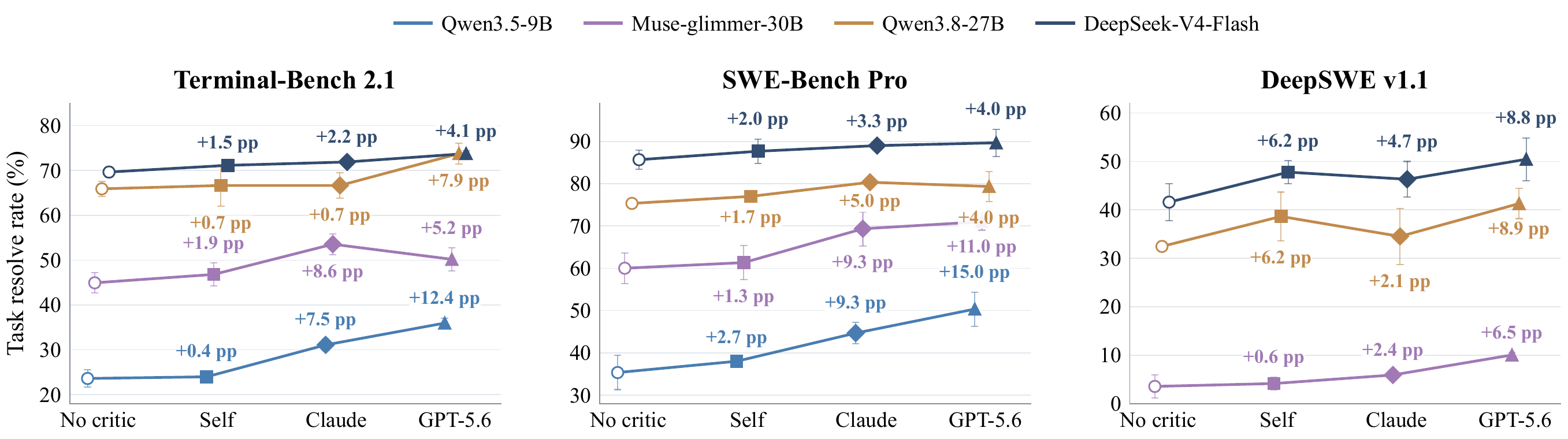}
    \caption{Task resolve rate with different critic models: the policy model itself (Self), Claude, and GPT. \model{} has consistent improvement regardless of the critic. Self-critique is effective where the policy is competent, while stronger critics bring large gains where the policy struggles.}
    \label{fig:critic_choice}
    \vspace{-10pt}
\end{figure}

We vary the critic model while keeping the policy fixed, using the policy model itself (Self), Claude, or GPT as the critic (\autoref{fig:critic_choice}). We exclude Qwen3.5-9B on DeepSWE, where the policy resolves no task under any critic (Section~\ref{sec:setup}). \model{} is robust to the choice of critic: all three critics improve over the no-critic agent in all but one of the remaining 33 policy--benchmark--critic combinations. Stronger critics generally yield larger gains, but no single critic is best in every setting. Notably, a policy model can serve as an effective critic of itself. On DeepSWE, self-critique improves Qwen3.8-27B and DeepSeek-V4-Flash by 6.2 pp each, matching or exceeding the gains from Claude. Since the self-critic has no knowledge beyond the policy's own, these gains suggest that much of the benefit comes from the framework: timely review, typed diagnosis, and follow-up help the model apply knowledge it already has but fails to use during execution. Self-critique does add inference compute, however, which this comparison does not match. Self-critique is less effective when the policy lacks the capability to solve the task, as for Muse Glimmer-30B on DeepSWE or Qwen3.5-9B on Terminal-Bench 2.1, likely because the critic shares the policy's blind spots. In these low-capability settings, a stronger critic brings substantial gains. With GPT as the critic, Qwen3.5-9B improves at least 12.4 pp, whereas self-critique yields at most 2.7 pp. Similarly, Muse Glimmer-30B gains 11.0 pp on SWE-Bench Pro and nearly triples its resolve rate on DeepSWE. 

\begin{takeaways}
    \item \model{} shows consistent improvement when critic model varies.
    \item Use self-critique when the policy is competent; use a stronger critic when it struggles; no critic helps a policy that never reaches a viable solution.
\end{takeaways}

\label{sec:main_results}

\section{Role of Audit}
\label{app:audit_ablation}

To isolate the role of the audits, we remove both the admission and release audits while keeping all other components of \model{} unchanged, so that every note proposed by the critic is delivered and every proposed closure is accepted (\autoref{tab:audit_ablation}). Removing the audits reduces the gain on all three benchmarks. The drop is largest on Terminal-Bench 2.1, where the gain over the non-critic agent falls from 7.9 to 2.6 pp, losing about two thirds of the improvement, and on DeepSWE, where it falls from 8.9 to 4.5 pp, about half. The results suggest that the audit acts as a filter against these harmful interventions, which matters much on benchmarks where intermediate evidence is easy to misread and misguided corrections are costly.

\begin{table}[tb!]
    \centering
    \footnotesize
    \caption{Audit ablation with Qwen3.8-27B as the policy and GPT-5.6-Sol as the critic. Resolve rate (\%), mean $\pm$ std over three runs; parentheses show gains over the non-critic agent (pp). Bold: best mean per benchmark.}
    \label{tab:audit_ablation}
    \begin{tabular}{@{}lccc@{}}
        \toprule
        Variant & TB2.1 & SWE-Pro & DeepSWE \\
        \midrule
        Non-critic
        & $65.9 \pm 1.7$
        & $75.3 \pm 1.2$
        & $32.4 \pm 1.0$ \\
        \midrule
        \model{} (w/ audit)
        & $\mathbf{73.8 \pm 2.3}$ {\scriptsize(+7.9)}
        & $\mathbf{79.3 \pm 3.5}$ {\scriptsize(+4.0)}
        & $\mathbf{41.3 \pm 3.1}$ {\scriptsize(+8.9)} \\
        \model{} (w/o audit)
        & $68.5 \pm 2.2$ {\scriptsize(+2.6)}
        & $78.3 \pm 2.1$ {\scriptsize(+3.0)}
        & $36.9 \pm 1.8$ {\scriptsize(+4.5)} \\
        \bottomrule
    \end{tabular}
\end{table} 

\section{Conclusion}

We presented \model{}, a verbal critic framework for long-horizon coding agents that treats each correction as a persistent note and follows it until the diagnosed problem is resolved. As a test-time critic, \model{} improves the resolve rate of non-critic agents by up to 15.0 percentage points across three benchmarks and four policy models, and achieves the highest mean resolve rate among competitive critic baselines with Qwen3.8-27B as the policy. Beyond inference, \model{}-guided rollouts provide approximately on-policy training data: fine-tuning Qwen3.5-9B on them matches distillation from a stronger model on held-out repositories (+10.2 pp) while preserving its Terminal-Bench 2.1 performance. Our training results use supervised fine-tuning on a single model with one out-of-domain benchmark; scaling this recipe to larger datasets, iterating it with the trained model as the new policy, or combining critic feedback with reinforcement learning to internalize self-correction could be the future work.

\bibliography{iclr2027_conference}
\bibliographystyle{plainnat}

\newpage
\appendix

\clearpage
\setcounter{tocdepth}{2}
\tableofcontents
\clearpage

\section{Implementation details of \model}
\label{app:impl_details}

\model{} is deployed as a proxy between the agent and its model endpoint. From the agent's perspective, nothing changes: the harness sends each request as it would to a model server, and the proxy forwards it. This design lets \model{} work with any harness without modification, and confines the critic to a single channel of influence, namely inserting at most one message into a forwarded request. The critic itself never executes commands, accesses files, or runs tests. It sees only what the agent has sent, and if it fails for any reason, such as a timeout or an unparseable reply, the request is forwarded unchanged, so the critic can never block the agent.

\textbf{Hybrid Review: }Most requests pass through without review. A review is triggered when the agent attempts to finish, when it appears stuck after three repeated or three consecutive failed commands, or when it has taken six work actions without a successful edit; otherwise, the critic checks in every 5 agent turns (10 on the longer DeepSWE tasks), starting from turn 3 and at least 2 turns apart. At each review, the critic reads the task and the rollout so far, presented as untrusted data and bounded to its context window: the 30 newest messages are kept verbatim, long messages are shortened to their beginning and end, and older messages are elided. Unlike a stateless judge, the critic also sees the currently open note, its resolution criterion, and the history of earlier notes, which allows it to follow up on its own feedback rather than start afresh.

\textbf{Audited Note Management: }A proposed note must pass several checks before it reaches the agent. While a note is open, the critic may only update it or let the agent continue, so that corrections do not pile up. Notes that refer to hidden tests or evaluation material, or that repeat the same suggestion within 6 turns, are discarded. A readiness note is also withheld while the agent is still exploring unfamiliar code, measured as more than half of its recent observations covering new ground. A candidate that survives these checks goes to the admission audit, a separate call to the critic model that either accepts it or turns it into \textit{continue}; the audit never rewrites the proposal, and no confidence threshold is involved. Once accepted, a note is delivered as a single message stating the operator, the guidance, and the resolution criterion, and asking the agent to fold the correction into its plan rather than reset its work. The message is inserted before the agent's next turn and stays at that position in later requests, so the agent sees one piece of advice it received a few turns ago instead of a fresh instruction at every step. If the note is delivered when the agent tries to finish, it takes the place of the finish attempt. As the agent continues, later reviews may refine the guidance while keeping the criterion fixed, or propose closing the note; closure takes effect only if the release audit finds the criterion met on the visible evidence, at which point the message disappears from the agent's context. Every review, together with its trigger, decision, filter and audit verdicts, and delivery outcome. 

\subsection{Baselines and Adaptations}
\label{sec:baselines}

All baselines run in the same harness as \model{}: the critic sits behind the agent's model endpoint, sees the same bounded rollout and task statement. For process critic baselines, they share the same hybrid review with \model{}. 

\textbf{SWE-PRM \citep{gandhi2025sweprm}.} We reproduce it as: at each review, the critic sees the task and the eight most recent steps, checks them against the paper's taxonomy of twelve trajectory inefficiencies, and delivers its overall guidance unless it judges the task on track. Unlike the original, which is invoked only periodically, it also reviews at the shared event triggers.

\textbf{SWE-Search \citep{antoniades2024swesearch}.} We reproduce its value function, including the prompt, the $-100$ to $100$ reward scale, and the evaluation of the last executed action. Since a single trajectory has no search tree, the feedback that SWE-Search writes for an alternative branch is delivered to the current trajectory when the reward is negative.

\textbf{LLM-as-verifier \citep{kwok2026llm} and Agentic Rubrics \citep{raghavendra2026agenticrubrics}.} Both methods were designed to score finished candidates for test-time selection. We run them as in-flight outcome checks at readiness events, i.e., when the agent attempts to finish or idles after its last edit, and deliver their own statement of what is missing when the candidate falls below the acceptance threshold. LLM-as-verifier scores the candidate on a 1--20 scale and accepts scores of at least 14. The original method takes the expectation over the score token's log-probabilities; we do so when the serving API exposes them (vLLM-served critics) and otherwise use the sampled score, which is the case for the GPT-5.6 critic in \autoref{tab:critic_comparison}. Agentic Rubrics writes up to eight repository-grounded rubric items once per task, scores the candidate item by item, and delivers failed required items. 

\section{More details of training recipe} \label{app:training_recipe}

\subsection{Data Preprocessing}
\label{app:training}

Training tasks are drawn from the 270 training-split tasks of SWE-Bench Pro, and no repo from the 108 held-out tasks is used. We select tasks where error recovery matters and both sources can supply successful demonstrations: the base Qwen3.5-9B resolves the task at most once in three runs without a critic, at least one critic-guided run of Qwen3.5-9B resolves it, and Qwen3.8-27B resolves it without a critic. This yields 67 tasks shared by both sources. For the \model{}-guided source, we keep every resolved trajectory of Qwen3.5-9B on these tasks from critic runs, with medium reasoning effort and for the teacher source, we run Qwen3.8-27B at medium reasoning effort without a critic. Each agent turn becomes one sample: the context contains the system prompt, the task, and all earlier actions and observations, and the target is the turn's reasoning, visible text, and tool call. A turn with multiple tool calls is split into single-call samples. We remove turns with invalid or rejected tool calls when the agent immediately retried, and drop samples longer than 128k tokens rather than truncating them.
Since the trained model runs without a critic, critic notes are removed from all contexts. On the 4.9\% of turns where an audited note was delivered, its diagnosis is instead rewritten as the model's own reasoning and prepended to that turn's thought, so the student learns to perform the diagnosis itself rather than wait for it. Sentences in which the policy refers to the critic are removed from both targets and histories, and samples whose commands or observations still mention it are dropped.

\subsection{Training details}
\begin{figure}[tb!]
    \centering
    \includegraphics[width=0.95\linewidth]{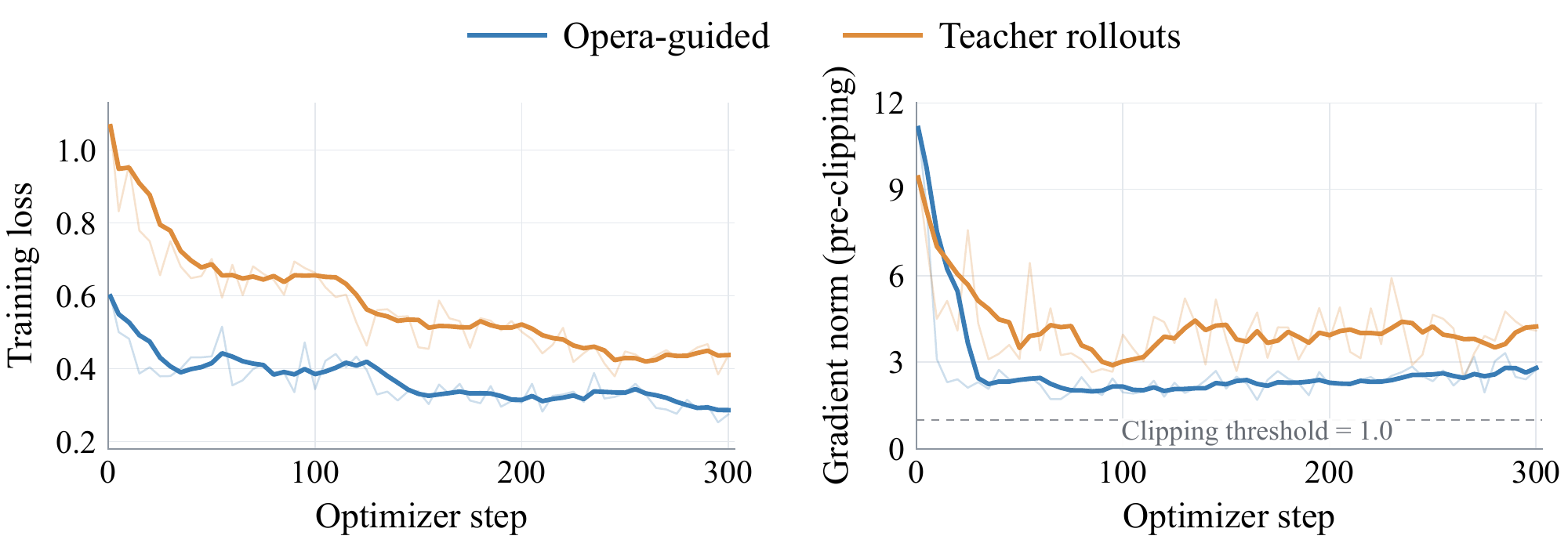}
    \caption{Training dynamics in SFT using \model{} guided self-reflection trajectories and Qwen3.8-27B trajectories. }
    \label{fig:training_dynamics}
\end{figure}

\autoref{tab:training_config} shows the hyperparameters we use during the SFT training and \autoref{fig:training_dynamics} shows the training dynamics (loss and gradient) in the two settings. 

\begin{table}[tb!]
    \centering
    \small
    \setlength{\tabcolsep}{6pt}
    \renewcommand{\arraystretch}{1.12}
    \caption{Training configuration. Shared settings apply to both trajectory sources; per-source statistics are listed at the bottom.}
    \label{tab:training_config}
        \begin{tabular}{@{}lcc@{}}
        \toprule
        \multicolumn{3}{@{}l}{\emph{Shared settings}} \\
        Student model & \multicolumn{2}{c}{Qwen3.5-9B, full-parameter, bf16} \\
        Framework & \multicolumn{2}{c}{ms-swift, DeepSpeed ZeRO-3} \\
        Maximum sequence length & \multicolumn{2}{c}{128k tokens} \\
        Loss & \multicolumn{2}{c}{Target turn only} \\
        Optimizer & \multicolumn{2}{c}{AdamW, $\beta_2{=}0.95$, wd 0.1, clip 1.0} \\
        Learning rate & \multicolumn{2}{c}{$2\times10^{-6}$, cosine, 3\% warm-up} \\
        Per-device batch / accumulation & \multicolumn{2}{c}{1 / 8} \\
        Epochs / seed & \multicolumn{2}{c}{3 / 42} \\
        \midrule
        \emph{Per source} & \model{}-guided & Qwen3.8-27B \\
        \midrule
        Tasks & 67 & 67 \\
        Training samples & 8{,}378 & 3{,}476 \\
        Supervised tokens & 2.69M & 2.25M \\
        GPUs (H100) & 8 & 4 \\
        Effective batch size & 64 & 32 \\
        Optimizer steps & 393 & 326 \\
        \bottomrule
    \end{tabular}
\end{table}

\section{Additional analysis of operator behavior and utility}

\subsection{Behavior responses to the different operators}

\begin{figure}[tb!]
    \centering
    \includegraphics[width=0.95\linewidth]{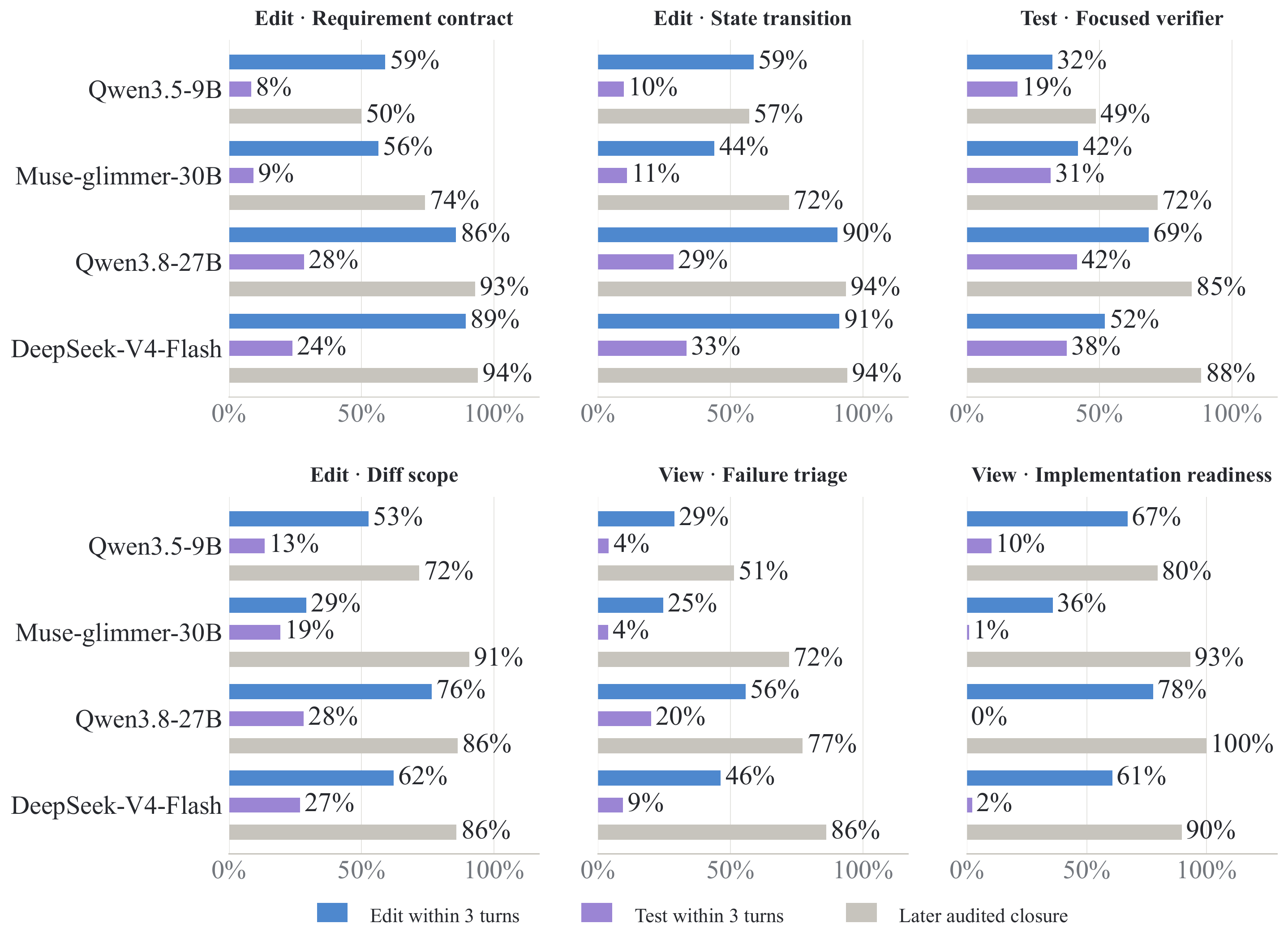}
    \caption{Agent responses to common operators. Bars show the percentages of delivered notes followed by an edit or a test within three turns, and the percentage eventually closed under audit.}
    \label{fig:operator_response}
\end{figure}

\begin{figure}[tb!]
    \centering
    \includegraphics[width=\linewidth]{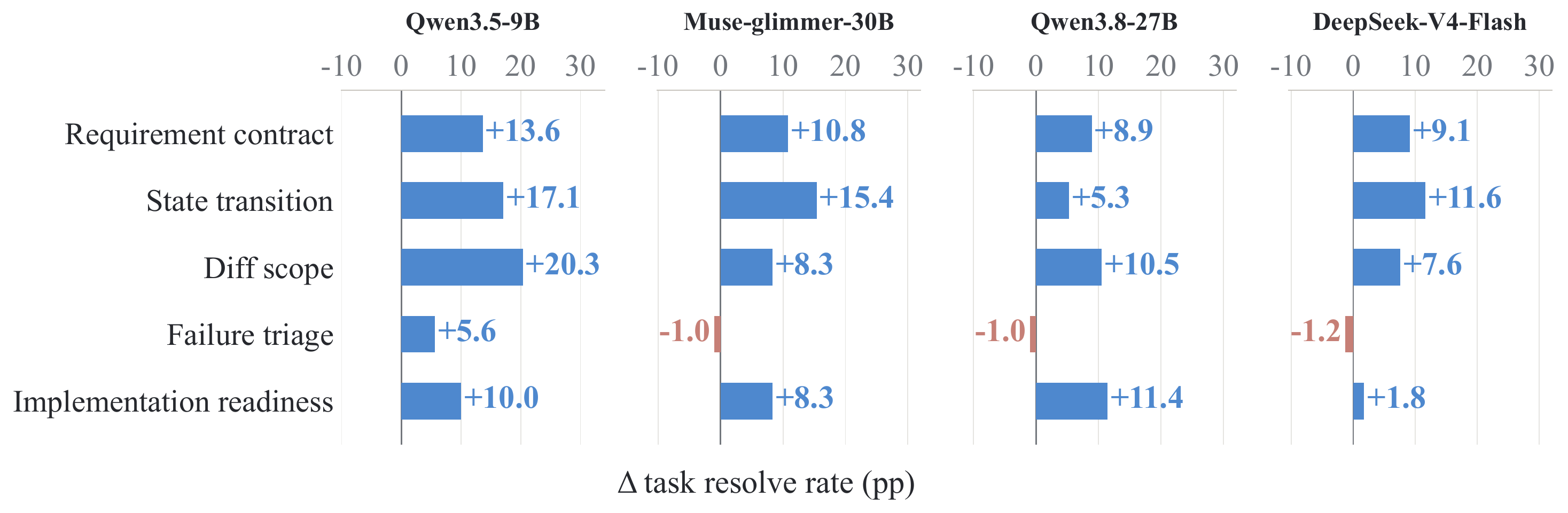}
    \caption{Task-success gains relative to baseline expectations, grouped by the first admitted operator. Common operator groups are shown. Gains reflect subsequent execution as a whole, not the isolated effect of the first operator.}
    \label{fig:first_operator_gain}
\end{figure}

\begin{figure}[tb!]
    \centering
    \includegraphics[width=\linewidth]{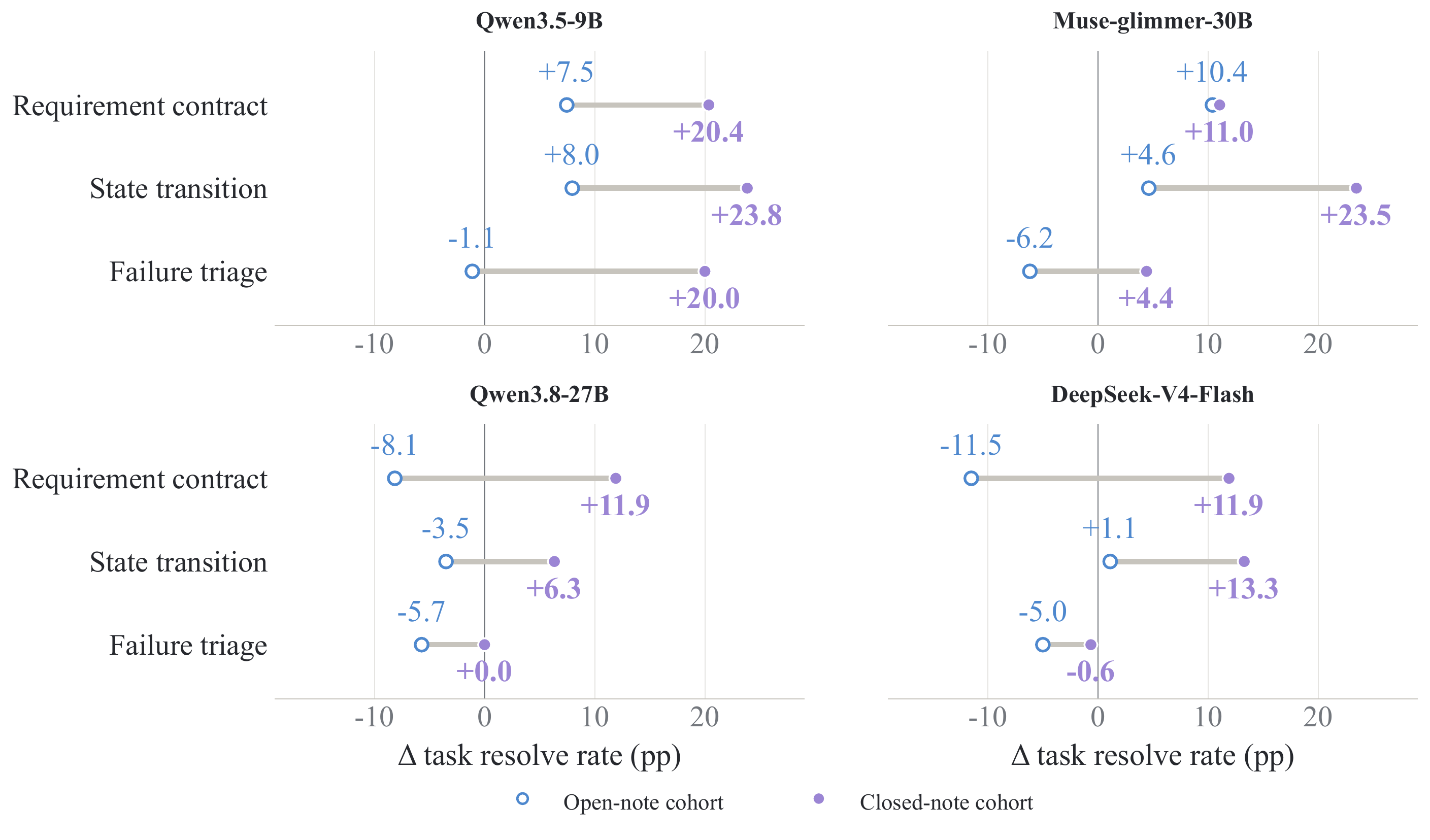}
    \caption{Task resolve-rate gains (pp) relative to no-critic expectations, grouped by policy model, first admitted operator, and note-closure status. Differences between open-note and closed-note cohorts describe associations rather than the causal effect of closure.}
    \label{fig:closure_conditioned}
\end{figure}

\textbf{Following feedback is not the same as resolving the issue.}
Stronger models act on feedback almost immediately and close most notes (\autoref{fig:operator_response}). Qwen3.5-9B edits after feedback as often as Muse-glimmer-30B but resolves only about half of its requirement-contract and state-transition notes, whereas Muse-glimmer-30B rarely edits right away yet eventually resolves most issues. Immediate adherence is thus a poor proxy for correction. Across models, agents seldom test unprompted, with the focused-verifier operator being the main trigger for testing, and failure-triage notes have the lowest edit and closure rates.

\textbf{Diagnoses that name a code change pay off; reinterpreting an error rarely does.}
Grouped by the first admitted operator, all three Edit-stage operators yield positive gains for every model (\autoref{fig:first_operator_gain}). Failure triage helps only Qwen3.5-9B and yields slightly negative gains for the other models, mirroring its low closure rates: correcting how an agent reads an error does not by itself tell it what to change.

\subsection{Critic behavior of using different critic models}

\begin{figure}[tb!]
    \centering
    \includegraphics[width=\linewidth]{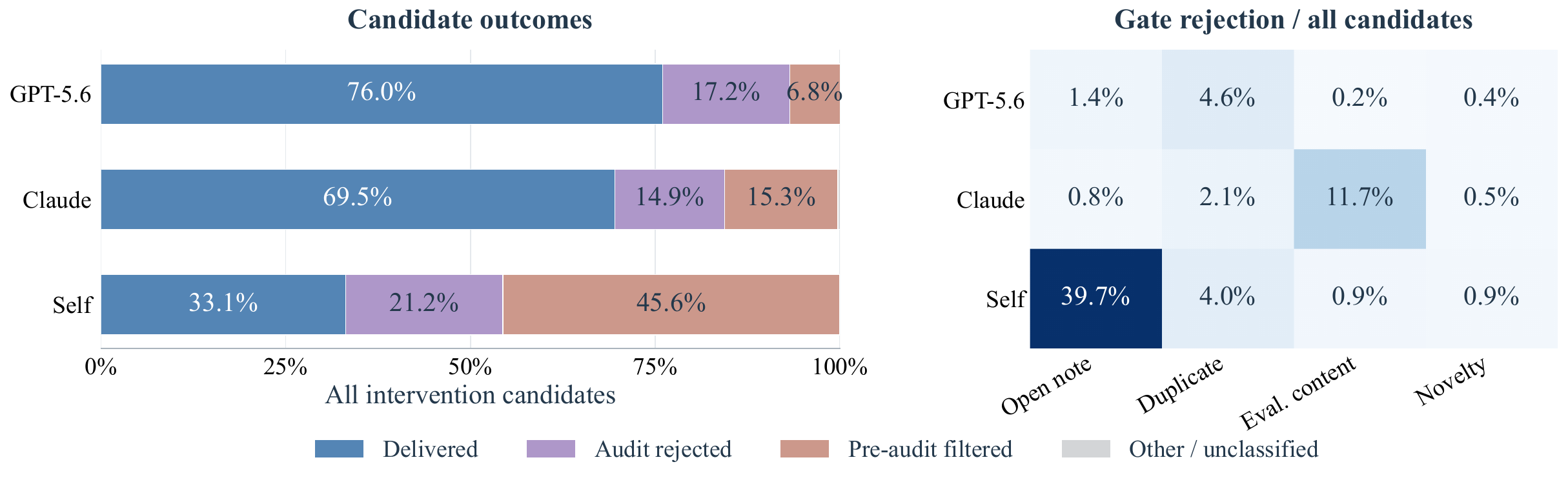}
    \caption{Intervention candidate outcomes across critic models. Left: delivered, audit-rejected, pre-audit-filtered, and unclassified candidates. Right: rejection rates for individual pre-audit gates. All percentages use all intervention candidates for the corresponding critic as the denominator.}
    \label{fig:candidate_outcome}
\end{figure}

\begin{figure}[tb!]
    \centering
    \includegraphics[width=\linewidth]{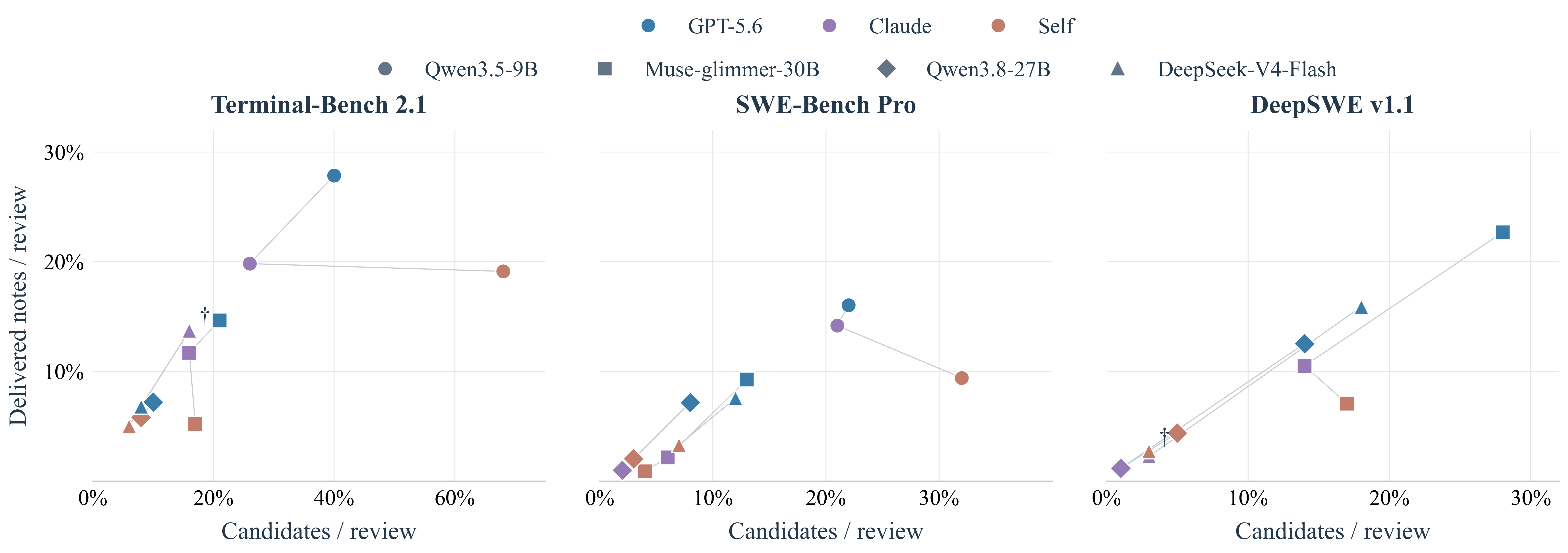}
    \caption{Proposal and delivery rates across critic and policy models. Each point represents a policy--critic setting within a benchmark: candidates per review on the x-axis and delivered notes per review on the y-axis. Colors denote critics, shapes denote policies, and lines connect the same policy across critics. Delivery rates are estimated from rounded aggregate statistics.}
    \label{fig:props_vs_deli}
\end{figure}

\textbf{The critic intervenes selectively, and more often for weaker policies.}
Most reviews end in \texttt{continue}: in nearly all settings, fewer than a third of reviews produce a candidate (\autoref{fig:props_vs_deli}). Intervention frequency follows the policy's needs: Qwen3.5-9B draws the most interventions on Terminal-Bench 2.1 and SWE-Bench Pro, the stronger Qwen3.8-27B and DeepSeek-V4-Flash far fewer, and all policies receive more on the harder DeepSWE. Among critics, GPT-5.6 generally delivers the most notes, consistent with its larger gains (Section~\ref{sec:sensitivity_to_critic}).

\textbf{Weaker critics rely more on the framework's filters.}
With GPT-5.6, most candidates are delivered, whereas self-critique delivers only a third, with nearly half filtered before the audit (\autoref{fig:candidate_outcome}). The dominant filter is the open-note constraint: the self-critic keeps proposing new issues instead of following up on its open note. For Qwen3.5-9B on Terminal-Bench 2.1, it proposes candidates at more than twice Claude's rate yet delivers notes at a similar rate. Filtering thus brings a weak critic's intervention frequency close to a stronger one's, which helps explain why self-critique remains effective. Stronger critics fail differently: Claude is filtered mainly for referring to evaluation material the agent cannot access.

\section{Case studies}
Case titles use the operator stages defined in Section \ref{sec:method_operators}; these need not match the action the agent was performing when the issue was detected.

\begin{operatorcase}{(1) Successful correction (search stage) --- \texttt{gravitational/teleport-b1bcd8b9}}
    \CaseField{Task}{The ingress reporter counts every HTTP connection as authenticated. Only TLS connections that present a client certificate may be counted; the fix has to live in the HTTP connection-state tracker (Active / Closed / Hijacked events).}
    \CaseField{Observation}{A dead search loop: \CaseProblem{eight consecutive \texttt{grep -rn ... --include='*.go' 2>/dev/null} calls returned nothing} because BusyBox grep does not support \texttt{--include} (it had printed \texttt{unrecognized option: include=*.go} once), and the agent hid that error with \texttt{2>/dev/null} and kept re-issuing the same form. After 20 turns the connection-state callback was still unlocated.}
    \CaseField{Operator}{\textit{repo\_localization\_search} for investigation that repeats a search form already shown to be unsupported.}
    \CaseField{Resolution criterion}{The Go code that registers the \texttt{http.Server.ConnState} callback is located, with visible handling of the Active, Closed and Hijacked events.}

    \CaseStep{T1}{\textbf{Agent.} Turns 13--21: \CaseProblem{empty-output searches for \texttt{metric}, \texttt{StateActive}, \ldots\ with the same unsupported flag}; in between it reads unrelated client code (\texttt{api/client/webclient/webclient.go}).}
    \CaseStep{T2}{\textbf{Critic.} Evidence: the empty results are an artefact of the suppressed error, not of the repository; the governing tracker remains unidentified. \CaseKey{Correction: run \texttt{git grep -nE 'ConnState|StateActive|StateClosed|StateHijacked' -- '*.go'} (pathspec-aware) and follow the callback to its tracker/reporter.} Admission audit: accepted (``repeated unsupported searches about eight times without progress''); note \texttt{F-aaa5aa2a} opened.}
    \CaseStep{T3}{\textbf{Agent.} Turn 22: runs the suggested \texttt{git grep} verbatim, which returns \texttt{lib/srv/ingress/reporter.go:89 HTTPConnStateReporter}; turns 24--27: views \texttt{reporter.go} and its test file and records that \texttt{StateNew} calls \texttt{ConnectionAuthenticated} without checking for TLS client certificates.}
    \CaseStep{T4}{\textbf{Follow-up.} Audited closure at turn 33: \CaseEvidence{the \texttt{git grep} output identified \texttt{HTTPConnStateReporter}, and the viewed implementation shows its \texttt{http.ConnState} callback handling \texttt{StateClosed} and \texttt{StateHijacked}} --- the exact site where \texttt{StateActive} handling and the TLS check must be added. Two later \textit{state\_transition\_review} notes (an uninitialised \texttt{tracker} in \texttt{NewReporter}; a close path that re-checked TLS instead of decrementing) were also closed, and \CaseEvidence{\texttt{go test ./lib/srv/ingress -run TestIngressReporter} passes at turn 66}.}

    \CaseField{Outcome}{\textbf{Note:} closed.
    \textbf{Task:} success (hidden tests pass; 0/3 without the critic).}
    \CaseField{Takeaway}{The agent was not wrong about \emph{what} to look for, only stuck in a tool form that silently returned nothing. One search-stage note replaced the loop with a single working query; localisation then unlocked the edit/test phase that the no-critic runs never reached.}
\end{operatorcase}

\begin{operatorcase}{(2) Successful correction (view stage) --- \texttt{protonmail/webclients-d3e51304}}
    \CaseField{Task}{Add \texttt{@proton/metrics} as a dependency and implement metric helpers: \texttt{getLabelID} must return \texttt{"custom"} for user-defined labels and the original ID for built-in mailboxes; \texttt{getPageSizeString} must normalise the page-size setting.}
    \CaseField{Observation}{A mis-read of the definition it had just viewed: \CaseProblem{after reading the \texttt{MAILBOX\_LABEL\_IDS} enum the agent wrote \texttt{import type \{ MAILBOX\_LABEL\_IDS \}} and the predicate \texttt{labelID in MAILBOX\_LABEL\_IDS}}. The \texttt{in} operator tests enum \emph{keys} (\texttt{INBOX}) while the mailbox IDs are the enum \emph{values} (\texttt{"0"}), and a type-only import leaves no runtime object to test against --- so the built-in-label branch can never be taken. The agent then treated the helper as finished and moved on to \texttt{package.json}.}
    \CaseField{Operator}{\textit{state\_transition\_review} for a branch that the required inputs can never reach.}
    \CaseField{Resolution criterion}{\texttt{getLabelID("0")} returns \texttt{"0"} while a user-defined ID returns \texttt{"custom"}.}

    \CaseStep{T1}{\textbf{Agent.} Turns 22--26: creates \texttt{mailMetricsHelper.ts} with \CaseProblem{the type-only import and the \texttt{in} predicate}, then edits \texttt{package.json}; nothing has exercised the helper.}
    \CaseStep{T2}{\textbf{Critic.} Evidence: \texttt{getLabelID("0")} falls into the \texttt{custom} branch (or fails at runtime because the enum is imported as a type). \CaseKey{Correction: import the enum as a value and classify by membership among \texttt{Object.values(MAILBOX\_LABEL\_IDS)}; verify by calling the helper and expecting \texttt{0 custom}.} Admission audit: accepted; note \texttt{F-415d225d} opened.}
    \CaseStep{T3}{\textbf{Agent.} Turn 27 (thought): ``the critic is correct''; turn 28: switches to a runtime import; turn 29: rewrites the predicate as value membership; turn 31: checks the logic with a \CaseProblem{separately built \texttt{VALUES} array instead of the exported function}. The critic updates the note (turn 33, \textit{minimal\_repro\_or\_focused\_verifier}): \CaseKey{call the exported \texttt{getLabelID} itself}; turn 33: the agent does so for \texttt{"0"} and \texttt{"custom-id"}.}
    \CaseStep{T4}{\textbf{Follow-up.} \CaseEvidence{The direct call prints \texttt{0} and \texttt{custom} with exit code 0}; audited closure at turn 37 cites exactly that output. (A third note on the same file is case (5).)}

    \CaseField{Outcome}{\textbf{Note:} closed.
    \textbf{Task:} success (0/3 without the critic).}
    \CaseField{Takeaway}{The agent had looked at the right definition but mis-read what the language construct tests. The note named the concrete input that breaks (\texttt{"0"}) and the observable to produce; the update then refused a proxy verification and insisted on the public interface, which is what the hidden tests call.}
\end{operatorcase}

\begin{operatorcase}{(3) Successful correction (edit stage) --- \texttt{qutebrowser-0b621cb0}}
    \CaseField{Task}{When a process cannot start, \texttt{GUIProcess.\_on\_error} must report the capitalised process name, the exact command in single quotes and the error kind (``failed to start:'', ``crashed:'', \ldots), and on non-Windows platforms append a hint that the command must exist and be executable.}
    \CaseField{Observation}{A wrong-branch edit: \CaseProblem{the first edit attached the new ``failed to start'' wording to \texttt{if error == QProcess.WriteError:}}, whereas a nonexistent command is reported by Qt as \texttt{QProcess.FailedToStart} and would still reach the old generic message. The task's own reproduction (start a nonexistent command) would therefore show none of the required wording; no test had been run yet.}
    \CaseField{Operator}{\textit{requirement\_contract\_review} for an edit that satisfies the wording but not the stated trigger.}
    \CaseField{Resolution criterion}{A nonexistent command reported as \texttt{FailedToStart} yields \texttt{Name 'cmd' failed to start: <detail>} followed by the non-Windows hint.}

    \CaseStep{T1}{\textbf{Agent.} Turn 5: views \texttt{\_on\_error}; turn 6: \CaseProblem{rewrites it so that the \texttt{WriteError} case produces the failed-to-start message, leaving \texttt{FailedToStart} to the generic branch}.}
    \CaseStep{T2}{\textbf{Critic.} Evidence: line 87 keys the message on \texttt{WriteError}; a real \texttt{FailedToStart} reaches lines 110--111. \CaseKey{Correction: move the formatting and the conditional non-Windows hint to the \texttt{FailedToStart} case and keep \texttt{WriteError} as its own message; verify with \texttt{pytest tests/unit/misc/test\_guiprocess.py}.} Admission audit: accepted (``the agent is mistakenly linking it to WriteError''); note \texttt{F-8349cefc} opened.}
    \CaseStep{T3}{\textbf{Agent.} Turn 7 (thought): ``the critic is correct''; turns 8--15: enumerates \texttt{QProcess.ProcessError} members (\texttt{FailedToStart = 0}); turn 16: re-edits the handler to key on \texttt{FailedToStart}; turns 20--22: finds and runs the unit test, whose log shows the message \CaseProblem{still lacks the hint}. Turn 28 update: \CaseKey{the hint predicate compares the whole detail to \texttt{"no such file or directory"} while Qt returns \texttt{execvp: No such file or directory} --- match the OS error text within the detail}; turn 30: the agent switches to a substring match; turn 31: reruns the test file.}
    \CaseStep{T4}{\textbf{Follow-up.} Audited closure at turn 35: \CaseEvidence{the visible pytest run started a real nonexistent command and logged \texttt{Testprocess 'this\_does\_not\_exist\_either' failed to start: execvp: No such file or directory (Hint: Make sure '...' exists and is executable)}}, satisfying the criterion even though the pre-existing test still asserts the obsolete format.}

    \CaseField{Outcome}{\textbf{Note:} closed.
    \textbf{Task:} success (0/3 without the critic).}
    \CaseField{Takeaway}{An edit-stage contract check caught a wrong-branch edit before any test had run, and the note's update handled the second-order mismatch (exact vs.\ substring) that the agent's own reasoning had missed; the no-critic runs ended with the same first edit in place.}
\end{operatorcase}

\begin{operatorcase}{(4) Successful correction (test stage) --- \texttt{qutebrowser-ef5ba1a0}}
    \CaseField{Task}{Locale override for QtWebEngine 5.15.3 on Linux: \texttt{\_get\_lang\_override(webengine\_version, locale\_name)} must return no override when the setting is off, the platform/version does not match, or the original \texttt{.pak} exists; a mapped fallback locale when its \texttt{.pak} exists; otherwise \texttt{en-US}.}
    \CaseField{Observation}{A green test run that proved nothing about the missing branch: \CaseProblem{the implementation built \texttt{locales\_dir} and went straight to the two \texttt{.pak} checks, so an unavailable directory falls through to \texttt{return 'en-US'} instead of returning \texttt{None}}; the agent then ran \texttt{tests/unit/config/test\_qtargs.py}, which does not cover that state, and read the result as confirmation.}
    \CaseField{Operator}{\textit{state\_transition\_review} for a final-default transition taken where an early no-override return is required.}
    \CaseField{Resolution criterion}{When the resolved \texttt{qtwebengine\_locales} directory is unavailable, \texttt{\_get\_lang\_override} returns \texttt{None}, not \texttt{en-US}.}

    \CaseStep{T1}{\textbf{Agent.} Turns 30--31: reads the existing test file and runs it; \CaseProblem{the unavailable-directory path is neither guarded in the code nor exercised by the tests}.}
    \CaseStep{T2}{\textbf{Critic.} Evidence: lines 377--398 construct the directory, check two \texttt{.pak} files and return \texttt{'en-US'}, with no availability guard. \CaseKey{Correction: return \texttt{None} immediately after resolving \texttt{locales\_dir} when it is unavailable; verify with a self-contained snippet that points \texttt{QLibraryInfo} at an empty temp dir and asserts \texttt{None}.} Admission audit: accepted; note \texttt{F-feda8c3f} opened.}
    \CaseStep{T3}{\textbf{Agent.} Turn 32 (thought) restates the requirement; turn 33: inserts \texttt{if not locales\_dir.exists(): return None} before the \texttt{.pak} checks; turn 34: runs the critic's snippet.}
    \CaseStep{T4}{\textbf{Follow-up.} \CaseEvidence{The snippet prints \texttt{Result: None} / \texttt{unavailable directory: no override}}; audited closure at turn 36. Two later notes on the same trial were also closed: \textit{diff\_scope\_review} (the \texttt{qt.workarounds.locale} option was read but never registered in \texttt{configdata.yml}; closed at turn 52 after \CaseEvidence{\texttt{registered}}) and \textit{requirement\_contract\_review} (the function took a \texttt{WebEngineVersions} object instead of the version value; closed at turn 63 after \CaseEvidence{\texttt{version accepted}}).}

    \CaseField{Outcome}{\textbf{Note:} closed.
    \textbf{Task:} success (0/3 without the critic).}
    \CaseField{Takeaway}{The agent's own test run was green for the branches it exercised; the note supplied the missing state (directory absent) together with a runnable check, turning an untested transition into an observable one.}
\end{operatorcase}

\begin{operatorcase}{(5) Successful correction (test stage, failure triage) --- \texttt{protonmail/webclients-d3e51304}}
    \CaseField{Task}{As in case (2): \texttt{getLabelID(labelID: string)} must accept any string and type-check.}
    \CaseField{Observation}{Half of a failure signature acted on, the other half abandoned: \CaseProblem{\texttt{tsc} reported TS2322 and TS2345 in the helper; the agent fixed the return statement (TS2322), then drifted into \texttt{grep} searches of the constants file while the second error, \texttt{TS2345: Argument of type 'string' is not assignable to parameter of type 'MAILBOX\_LABEL\_IDS'} at \texttt{includes(labelID)}, was still printed by the re-run}.}
    \CaseField{Operator}{\textit{failure\_signature\_triage} for a visible compiler error that the latest actions no longer address.}
    \CaseField{Resolution criterion}{\texttt{mailMetricsHelper.ts} type-checks with \texttt{getLabelID} accepting arbitrary strings for the enum-value membership test.}

    \CaseStep{T1}{\textbf{Agent.} Turn 37: \texttt{tsc --noEmit} prints both errors; turn 38: edits the return statement; turns 43--45: \CaseProblem{greps the enum definition and re-runs \texttt{tsc}, which still prints TS2345 at line 36}.}
    \CaseStep{T2}{\textbf{Critic.} Evidence: \texttt{Object.values(MAILBOX\_LABEL\_IDS)} infers an enum-typed array, so \texttt{.includes(labelID)} rejects a plain \texttt{string}. \CaseKey{Correction: type the collected values as \texttt{readonly string[]} while keeping the declared \texttt{MAILBOX\_LABEL\_IDS | 'custom'} return; re-run \texttt{tsc} and expect no diagnostic for the helper file.} Admission audit: accepted; note \texttt{F-2cd990db} opened.}
    \CaseStep{T3}{\textbf{Agent.} Turn 46: declares \texttt{const MAILBOX\_LABEL\_ID\_VALUES:\allowbreak\ readonly string[] =\allowbreak\ Object.values(...)}; turn 47: re-runs \texttt{tsc} filtered to the helper file.}
    \CaseStep{T4}{\textbf{Follow-up.} \CaseEvidence{The filtered \texttt{tsc} output is empty (the 31 remaining errors are all in other files)}; audited closure at turn 51. Turn 50 then exercises \texttt{getPageSizeString} and \texttt{getLabelID} end to end before the agent finishes.}

    \CaseField{Outcome}{\textbf{Note:} closed.
    \textbf{Task:} success.}
    \CaseField{Takeaway}{The failure signature was already on screen; the agent had acted on half of it and then started searching. The triage note re-attached the remaining error line to a one-line fix and a re-run, instead of letting the search continue.}
\end{operatorcase}

\begin{operatorcase}{(6) Successful correction (submit stage) --- \texttt{qutebrowser-8cd06741}}
    \CaseField{Task}{Expose the QtWebEngine 6.6 dark-mode image classifier: \texttt{colors.\allowbreak webpage.\allowbreak darkmode.\allowbreak policy.\allowbreak images} gains the value \texttt{smart-simple}; on Qt 6.6+ \texttt{smart} must emit \texttt{ImagePolicy=2,\allowbreak\ ImageClassifierPolicy=0} and \texttt{smart-simple} must emit \texttt{ImageClassifierPolicy=\allowbreak 1}; older Qt versions must be unaffected.}
    \CaseField{Observation}{A completion without the central check: \CaseProblem{the one direct \texttt{darkmode.settings()} call had crashed with \texttt{AttributeError: 'NoneType' object has no attribute 'get'} (no config instance) and was never retried}; instead the agent inspected \texttt{\_MultiMappingSetting.tuples()} and the internal mapping tables, marked all five task-tracker items done and was about to call \texttt{finish}.}
    \CaseField{Operator}{\textit{submission\_readiness\_review} for verification that bypasses the interface the task is about.}
    \CaseField{Resolution criterion}{A direct \texttt{darkmode.settings()} call shows Qt 6.5 emitting only \texttt{ImagePolicy=2} for both values and Qt 6.6 additionally emitting \texttt{ImageClassifierPolicy} 0 or 1.}

    \CaseStep{T1}{\textbf{Agent.} Turns 102--106: \CaseProblem{updates the task tracker to five ``done'' items and prints \texttt{\_IMAGE\_POLICIES} and the classifier mapping as its final check}; the failed end-to-end attempt (turn 101) is not retried.}
    \CaseStep{T2}{\textbf{Critic.} Evidence: every successful check reads internals; none shows configured values flowing through \texttt{settings()} into the emitted switch data. \CaseKey{Correction: run one end-to-end script that mocks \texttt{config.instance.get} and asserts the four required \texttt{settings()} outputs for Qt 6.5 and 6.6; expect \texttt{settings() end-to-end OK}.} Admission audit: accepted (``bypasses the necessary interface''); note \texttt{F-b81e67aa} opened.}
    \CaseStep{T3}{\textbf{Agent.} Turn 108: runs the script verbatim.}
    \CaseStep{T4}{\textbf{Follow-up.} \CaseEvidence{Output \texttt{settings() end-to-end OK}} --- the close audit accepts on that output and the agent finishes on the next action (turn 109). Earlier in the same trial a \textit{requirement\_contract\_review} note (opened turn 43, three updates, closed turn 86) and a \textit{state\_transition\_review} note (a shallow \texttt{copy.copy} in \texttt{copy\_replace\_setting} mutated the shared \texttt{\_Setting} objects so building the Qt 6.4 definition changed Qt 5.15.3's keys; closed at turn 101 after \CaseEvidence{\texttt{Settings are independent!}}) had been closed.}

    \CaseField{Outcome}{\textbf{Note:} closed.
    \textbf{Task:} success (0/3 without the critic).}
    \CaseField{Takeaway}{The submit-stage note did not change the patch; it forced the one verification the agent had skipped after an unrelated setup error, and in this trial that verification passed. The same operator is what blocks premature \texttt{finish} calls when it does not.}
\end{operatorcase}

\end{document}